\documentclass[10pt,twocolumn,letterpaper]{article}

\usepackage[pagenumbers]{iccv} %

\definecolor{iccvblue}{rgb}{0.21,0.49,0.74}
\usepackage[pagebackref,breaklinks,colorlinks,allcolors=iccvblue]{hyperref}

\usepackage{graphicx}
\usepackage{array}
\usepackage{color,soul}
\usepackage{multirow}
\usepackage{wrapfig}
\usepackage{adjustbox}
\usepackage{titletoc}
\usepackage{algorithm}
\usepackage{algpseudocode}

\definecolor{lightblue}{rgb}{.90,.95,1}
\sethlcolor{lightblue}

\def\model{UIP2P} \title{\model{}: Unsupervised Instruction-based Image Editing \\via Edit Reversibility Constraint} 

\author{\vspace{1mm}
    Enis Simsar$^{1}$
    \hspace{0.5cm}
    Alessio Tonioni$^{3}$
    \hspace{0.5cm}
    Yongqin Xian$^{3}$
    \hspace{0.5cm}
    Thomas Hofmann$^{1}$
    \hspace{0.5cm}
    Federico Tombari$^{2,3}$
    \\
    $^1$ETH Zürich - DALAB
    \hspace{2.5em} 
    $^2$Technical University of Munich
    \hspace{2.5em} 
    $^3$Google Switzerland\\
    {\tt\small \href{https://uip2p.github.io}{https://uip2p.github.io}}
}
\begin{document}

\twocolumn[{

    \small        
    \renewcommand\twocolumn[1][]{#1}%
    \maketitle
    \captionsetup{type=figure}
    \vspace{-1.8em}  
    \includegraphics[width=\textwidth]{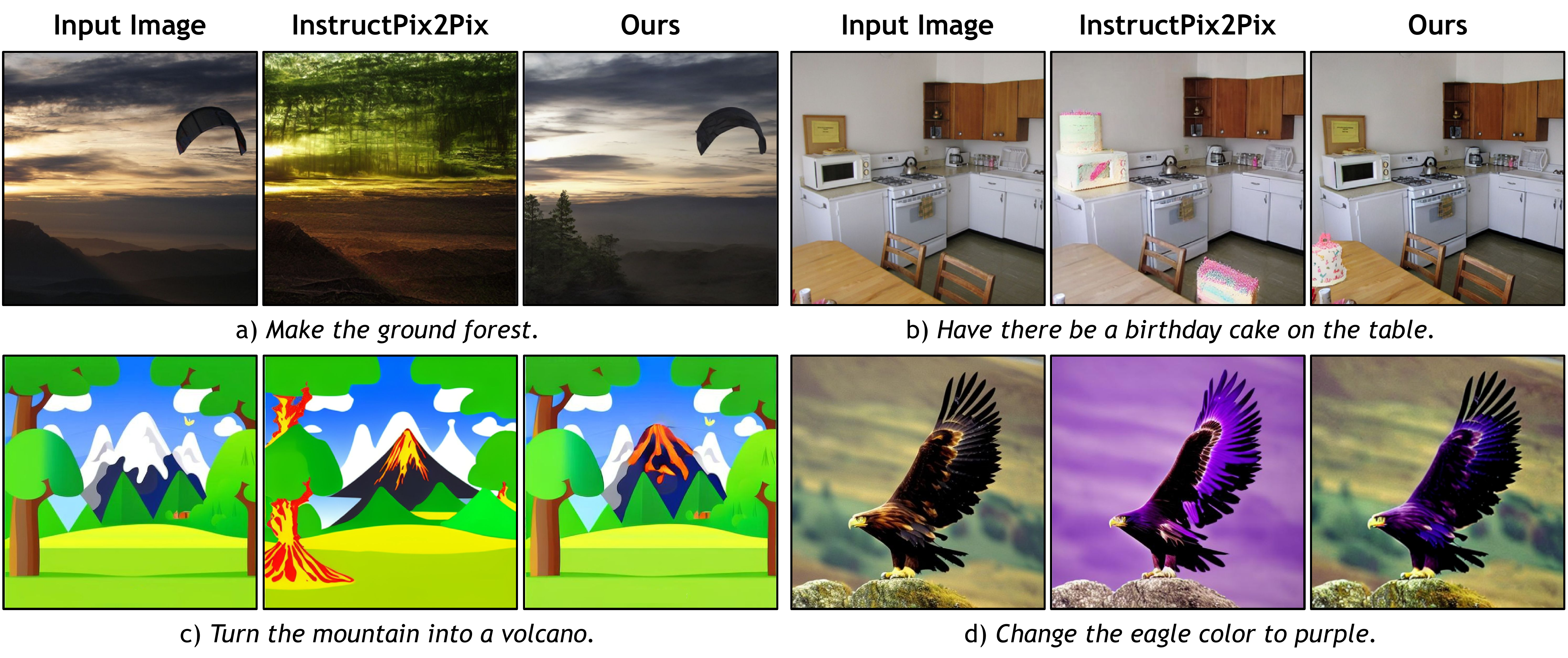}    
    \vspace{-1.5em}

    \captionof{figure}{\textbf{\underline{U}nsupervised \underline{I}nstruct\underline{P}ix\underline{2}\underline{P}ix.} Our approach applies more precise and coherent edits while better preserving scene structure. \model{} surpasses the supervised alternative, IP2P, trained on a synthetic dataset, demonstrating superior performance on both real (a, b) and synthetic (c, d) images.}
    \vspace{1.5em}
    \label{fig:teaser}
}]

\begin{abstract}
We propose an unsupervised instruction-based image editing approach that removes the need for ground-truth edited images during training. Existing methods rely on supervised learning with triplets of input images, ground-truth edited images, and edit instructions. These triplets are typically generated either by existing editing methods—introducing biases—or through human annotations, which are costly and limit generalization.
Our approach addresses these challenges by introducing a novel editing mechanism called Edit Reversibility Constraint (ERC), which applies forward and reverse edits in one training step and enforces alignment in image, text, and attention spaces. This allows us to bypass the need for ground-truth edited images and unlock training for the first time on datasets comprising either real image-caption pairs or image-caption-instruction triplets. 
We empirically show that our approach performs better across a broader range of edits with high-fidelity and precision. By eliminating the need for pre-existing datasets of triplets, reducing biases associated with current methods, and proposing ERC, our work represents a significant advancement in unblocking scaling of instruction-based image editing.
\end{abstract}

\begin{figure*}[!tb]
  \centering
  \includegraphics[width=0.95\textwidth]{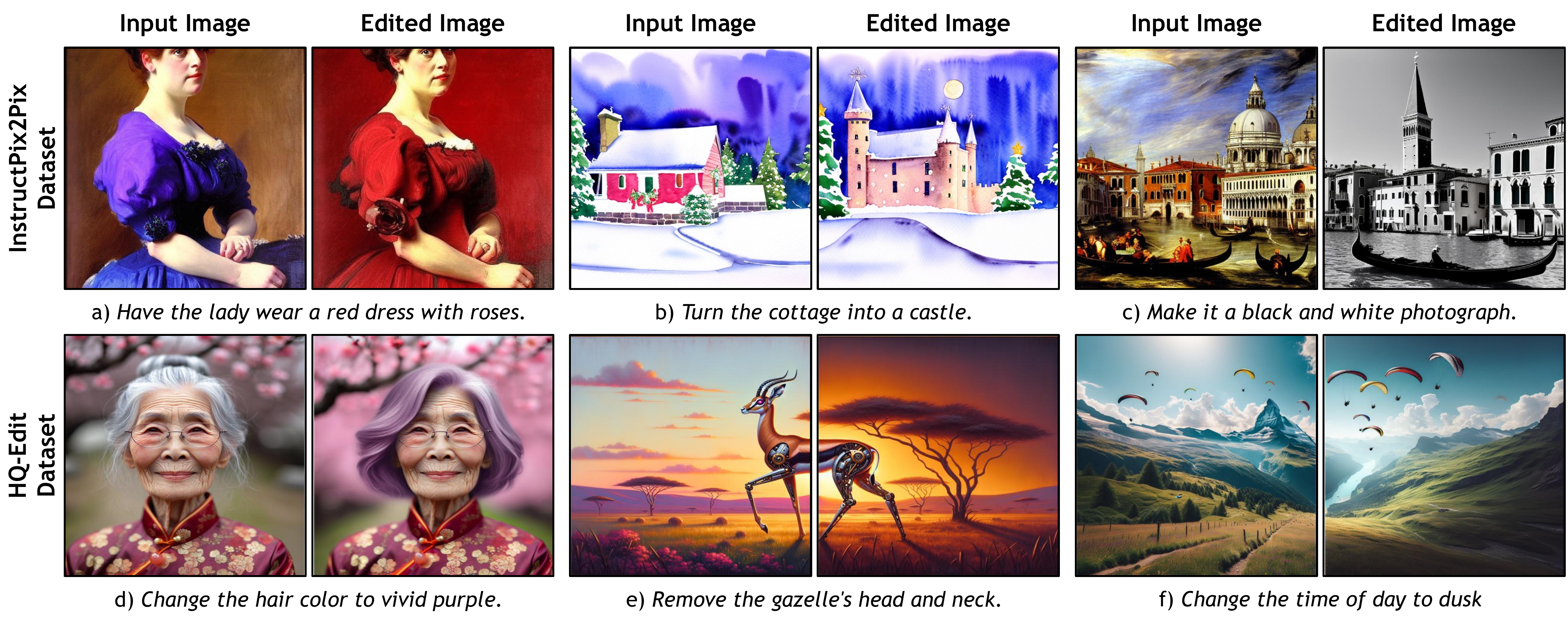}
  \vspace{-0.5em}
    \caption{\textbf{Supervision Biases in InstructPix2Pix and HQ-Edit datasets.} Each example shows an input image and its corresponding ground-truth edited image for the given edit instruction. InstructPix2Pix employs Prompt-to-Prompt, while HQ-Edit uses DALL-E 3 and GPT-4V. \textit{(a \& d) Attribute-entangled edits:} Modifying a specific feature, such as clothing or hair color, unintentionally alters surrounding textures or elements. \textit{(b \& e) Scene-entangled edits:} Transforming objects, like turning a cottage into a castle or removing an element, affects unintended parts of the scene. \textit{(c \& f) Global changes:} Edits like converting an image to black and white or changing the time of day introduce widespread scene modifications, often compromising visual preservation.} 
    \label{fig:ptp_entanglement}
\end{figure*}

\section{Introduction}
\label{sec:intro}
Diffusion models (DMs) have recently made significant advancements in generating high-quality and diverse images, driven largely by breakthroughs in text-to-image generation~\cite{ho2020denoising, Saharia2022Imagen, Rombach2022StableDiffusion, Ramesh2022DALLE2}. This progress has enabled various techniques for tasks such as personalized image generation~\cite{Ruiz2022DreamBooth, wei2023elite, gal2022image}, context-aware inpainting~\cite{lugmayr2022repaint, Nichol2022GLIDE, yang2023paint}, and image editing based on textual prompts~\cite{Avrahami2022BlendedDiffusion, Hertz2022Prompt2Prompt, Meng2022SDEdit, Kawar2022Imagic, couairon2023diffedit}. 
Beyond prompt-based editing, DMs have demonstrated strong capabilities in instruction-based editing, where edits are guided by explicit natural language commands, such as instructions~\cite{Brooks2022InstructPix2Pix, hui2025hqedit, Zhang2023MagicBrush, huang2024smartedit, Zhang2023HIVE, fu2023guiding}.

However, existing methods predominantly rely on supervised learning with ground-truth edited images, \ie{,} requiring datasets of triplets containing input images, edited images, and edit instructions~\cite{Brooks2022InstructPix2Pix, Zhang2023MagicBrush, Zhang2023HIVE, fu2023guiding, hui2025hqedit}. These datasets are often generated using existing editing methods, such as Prompt-to-Prompt~\cite{Hertz2022Prompt2Prompt}, or more recently, advanced diffusion and multimodal models like DALL-E 3~\cite{dalle3} and GPT-4V~\cite{achiam2023gptv}, as in the HQ-Edit dataset~\cite{hui2025hqedit}. While automated approaches enable scalable dataset creation, they introduce biases (see~\cref{fig:ptp_entanglement}), including (a \& d) attribute-entangled edits that unintentionally alter surrounding elements, (b \& e) scene-entangled edits, and (c \& f) global changes that significantly modify the entire scene. Conversely, while valuable, human-annotated data~\cite{Zhang2023MagicBrush} is impractical for large-scale training due to the high cost and effort involved. Reliance on automated or human-generated ground-truth edits limits the diversity of achievable modifications, constraining the development of models capable of accurately understanding and executing user instructions.

We present \model{}, an unsupervised training for instruction-based image editing that eliminates the reliance on triplet datasets, whether generated or human-annotated, by introducing the Edit Reversibility Constraint (ERC)—a constraint enforced while training through forward and reverse edits. 
Instead of relying on explicitly supervised edited images, we ensure that modifications remain precise and follow the given instructions by leveraging CLIP’s ability to align text and images in a shared embedding space~\cite{radford2021learning}. %
Additionally, we explicitly enforce alignment between forward and reverse edits in both the image and attention spaces, enabling \model{} to accurately interpret and localize user instructions while ensuring edits remain coherent and reflect the intended changes. ERC allows \model{} to preserve the integrity of the original content while making precise adjustments, further enhancing the reliability of the edits.
Some previous methods have enforced cycle consistency within predefined domain transformations~\cite{su2022dual,cyclediffusion,Zhu2017CycleGAN,xu2023cyclenet, kim2022diffusionclip}. In contrast, our constraint applies to any arbitrary pair of forward and reverse edits within a single training step, leveraging the expressive power of modern diffusion models to ensure flexibility across diverse editing tasks.
By eliminating the dependence on pre-existing datasets, our approach enables training on large-scale real-image datasets—a capability previously constrained by the limitations of existing methods and the high cost of human labeling. As a result, \model{} significantly expands the scope and scalability of instruction-based image editing compared to prior approaches.
Our key contributions are as follows:

\begin{itemize}
    \item We propose \model{}, an unsupervised training approach for instruction-based image editing that eliminates the need for ground-truth edited images. By doing so, \model{} offers a more scalable solution while reducing the biases inherent in alternative supervised methods.
    \item We introduce the Edit Reversibility Constraint (ERC), a novel mechanism that ensures edits remain stable when applying both forward and reverse edits, preserving structural integrity in the image, text and attention space. This enables precise, high-fidelity modifications that accurately reflect user instructions.
    \item Our approach demonstrates scalability and versatility across various real-image datasets, allowing a broad range of edits without relying on pre-existing datasets, significantly expanding the capabilities of instruction-based image editing.
\end{itemize}

\section{Related Work}
\label{sec:related_work}

\vspace{2pt}\noindent\textbf{CLIP-Based Image Manipulation.} StyleCLIP \cite{patashnik2021styleclip} combines StyleGAN and CLIP for text-driven image manipulation, requiring optimization for each specific edit. Similarly, StyleGAN-NADA \cite{Gal2022StyleGAN-NADA} enables zero-shot domain adaptation by using CLIP guidance to modify generative models. While these approaches allow for flexible edits, they often rely on domain-specific models or optimization processes for each new task. These works illustrate the potential of CLIP’s powerful semantic alignment for image manipulation, which motivates the use of CLIP in other generative frameworks, such as diffusion models.

\vspace{4pt}\noindent\textbf{Text-based Image Editing with Diffusion Models.} One common approach in image editing is to use pre-trained diffusion models by first inverting the input image into the latent space and then applying edits through text prompts~\cite{Mokady2022NullTextInversion, Hertz2022Prompt2Prompt, wang2023mdp, Meng2022SDEdit, couairon2023diffedit, ju2023direct, parmar2023zero, wang2023dynamic, wu2023uncovering}. For example, DirectInversion~\cite{ju2023direct} edits the image after inversion using Prompt-to-Prompt~\cite{Hertz2022Prompt2Prompt}, but the inversion step can lead to losing essential details from the original image. Additionally, methods like DiffusionCLIP \cite{kim2022diffusionclip}, CycleDiffusion~\cite{cyclediffusion}, CycleNet~\cite{xu2023cyclenet}, and DualDiffusion~\cite{su2022dual} explore \textit{domain-to-domain translation} as a way to propose image editing, by using textual prompt or without any conditions. However, their focus on translating between two fixed domains makes it difficult to handle complex edits, such as the insertion or deletion of objects. 
In contrast, our approach is designed for general-purpose instruction-based image editing without being restricted to domain translation, enabling greater flexibility in handling a broader range of modifications.

\vspace{4pt}\noindent\textbf{Instruction-based Image Editing with Diffusion Models.} Another line of methods for image editing involves training models on datasets containing triplets of input image, edit instruction, and edited image~\cite{Brooks2022InstructPix2Pix, Zhang2023MagicBrush, Zhang2023HIVE, hui2025hqedit}. These methods, since they directly take the input image as a condition, do not require an inversion step.
InstructDiffusion~\cite{geng2023instructdiffusion} builds on InstructPix2Pix by handling a wider range of vision tasks but has difficulty with more advanced reasoning. MGIE~\cite{fu2023guiding} improves on this by using large multimodal language models to generate more precise instructions. SmartEdit~\cite{huang2024smartedit} goes a step further by introducing a Bidirectional Interaction Module that better connects the image and text features, improving its performance in challenging editing scenarios. UltraEdit~\cite{zhao2024ultraedit} scales instruction-based editing to larger datasets and models, demonstrating improved performance on fine-grained editing tasks.

A major challenge in instruction-based image editing is the need for large-scale, high-quality triplet datasets. InstructPix2Pix~\cite{Brooks2022InstructPix2Pix} partially addresses this by generating extensive datasets using GPT-3~\cite{Brown2020GPT3} and Prompt-to-Prompt~\cite{Hertz2022Prompt2Prompt}. However, while this mitigates data scarcity, it introduces issues like model biases from Prompt-to-Prompt. MagicBrush~\cite{Zhang2023MagicBrush} tackles the quality aspect with human-annotated datasets, but this approach is small-scale, limiting its practicality for broader use.

Our method leverages CLIP's semantic space for aligning images and text, providing a more robust solution. Moreover, Edit Reversibility Constraint (ERC) tackles both dataset limitations and biases by enforcing coherence between forward and reverse edits. Our approach enhances scalability and precision for complex instructions and eliminates the dependency on triplet datasets, making it applicable to any image-caption dataset of real images. Moreover, as ERC modifies only the training objectives of InstructPix2Pix, it integrates seamlessly with any model extension.

\section{Background}
\label{sec:background}
Our method builds upon InstructPix2Pix (IP2P) \cite{Brooks2022InstructPix2Pix}, which uses Stable Diffusion's U-Net architecture \cite{Rombach2022StableDiffusion} for instruction-based image editing. IP2P conditions the model on both input images and text instructions using classifier-free guidance \cite{ho2021classifierfree}. Cross-attention mechanisms compute attention maps based on text conditioning, enabling localized edits by focusing on relevant image regions.

IP2P trains on triplets of input images, edit instructions, and edited images generated using Prompt-to-Prompt~\cite{Hertz2022Prompt2Prompt}. However, this reliance on synthetic data introduces limitations: (1) poor generalization to real-world images, and (2) inherited biases from Prompt-to-Prompt, as shown in~\cref{fig:ptp_entanglement}.

\begin{figure*}[!tb]
  \centering
  \includegraphics[width=.95\textwidth]{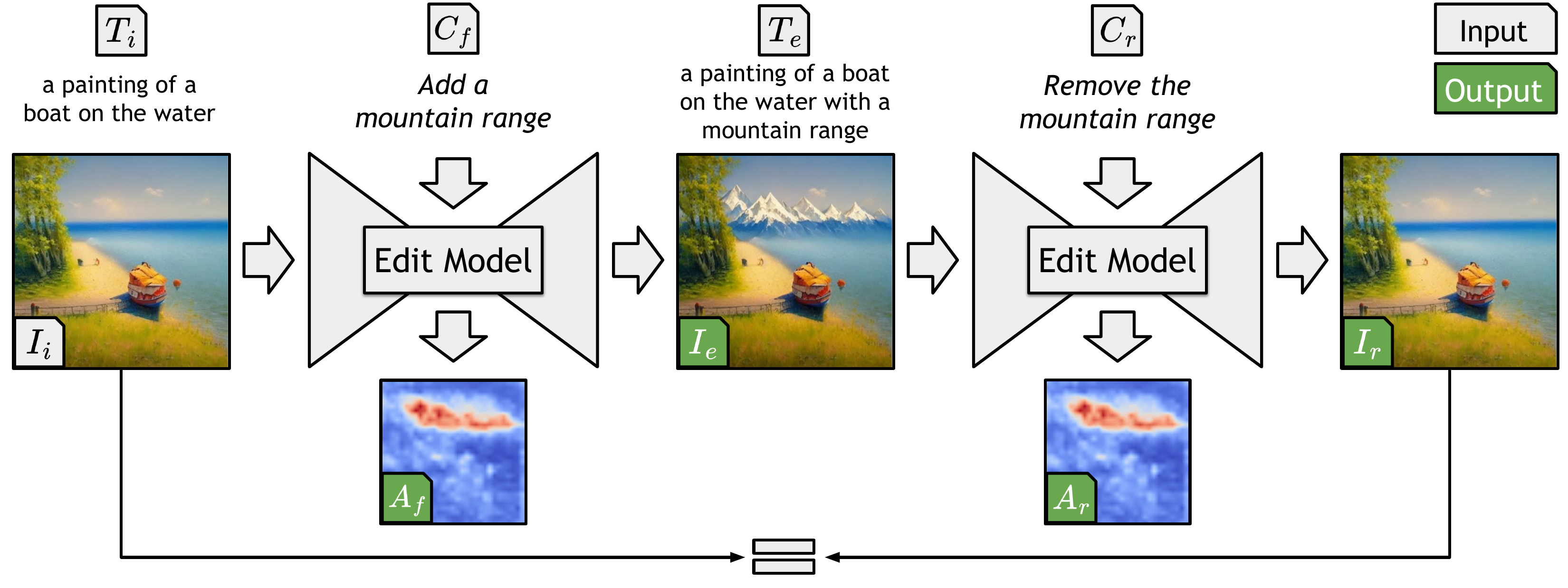}
    \caption{\textbf{Overview of the \model{} training framework.} The model learns instruction-based image editing by applying forward and reverse instructions. Starting with an input image and a forward instruction, shared \textit{Edit Model} generates an edited image. A reverse instruction is then applied to reconstruct the original image, enforcing Edit Reversibility Constraint (ERC).}
  \label{fig:method} %
\end{figure*}

\section{Method}
\label{sec:method}
We introduce \model{}, an unsupervised diffusion-based framework for instruction-based image editing, enforcing the Edit Reversibility Constraint (ERC) to ensure that applied edits can be undone using corresponding reverse instructions. Unlike prior methods reliant on synthetic triplets or domain-specific transformations, \model{} operates on real-world image-captions datasets without needing edit pairs. Our approach integrates text-image alignment, attention-based localization, reconstruction fidelity, and efficient noise prediction, enabling precise and generalizable edits. Additionally, we leverage LLM-generated reverse instructions, eliminating the need for requiring supervision while significantly improving scalability. 
In the following sections, we describe our approach in detail, including the key components of our framework (\cref{subsec:method}), the loss functions that enforce edit reversibility, and the training data generation procedure (\cref{subsec:data}).

\subsection{Framework}
\label{subsec:method}

\subsubsection{\model{}}
To enforce the Edit Reversibility Constraint (ERC), we define each training step as a sequence of a forward edit followed by its corresponding reverse edit, both applied using the same \textit{Edit Model}. This ensures that modifications can be undone accurately, embedding edit constraints directly within the training process, see \cref{fig:method} for an overview.
A key advantage of our approach is that training is performed \textit{in a single denoising step}, significantly reducing computational cost compared to existing methods that require multiple refinement passes per edit. 
To enforce the semantic correctness of ERC, we introduce three core components:

\begin{enumerate}
    \item \textbf{Text and Image Direction Alignment}: We leverage CLIP embeddings~\cite{Radford2021CLIP} to align the semantic relationship between textual instructions and image modifications without the need of ground-truth edited images. By operating within CLIP’s embedding space, our model ensures that changes in the input and edited images correspond to changes in their respective captions. This alignment is critical for enforcing the Edit Reversibility Constraint (ERC), ensuring that edits accurately follow instructions while preserving the input image’s structure.
    
    \item \textbf{Attention Map Alignment}: To ensure stable and well-localized edits, we enforce alignment between cross-attention maps generated during both forward and reverse edits. This ensures the model focuses on the same regions of the image when applying and undoing modifications. Attention Map Alignment also regularizes the training objective, improving spatial precision in edits.
    
    \item \textbf{Reconstruction Alignment}: A key aspect of ERC is the ability to reconstruct the original input after applying the reverse instruction. This ensures that the model reliably undoes its edits. We achieve this by minimizing both pixel-wise and semantic discrepancies between the reconstructed image and the original input, reinforcing the accuracy of the applied and reversed edits.

\end{enumerate}

\begin{table*}[!htb]
  \caption{\textbf{Reverse Instruction Generation.} Our method generates reverse instructions for the IP2P dataset, eliminating the need for ground-truth edited images. Additionally, edit instructions, edited captions, and reverse instructions are generated for CC3M and CC12M datasets—denoted as CCXM. The \hl{texts} are generated by LLMs such as GEMINI, and GEMMA2.}
  \label{tab:samples}
  \centering
  \resizebox{\textwidth}{!}{
    \begin{tabular}{@{}p{0.04\linewidth}@{\hskip 0.05in}>{\raggedright}p{0.3\linewidth}@{\hskip 0.05in}>{\raggedright}p{0.2\linewidth}@{\hskip 0.05in}>{\raggedright}p{0.3\linewidth}@{\hskip 0.05in}>{\raggedright\arraybackslash}p{0.2\linewidth}@{}}
        \toprule
         & Input Caption & Edit Instruction & Edited Caption & Reverse Instruction\\
        \midrule
        \multirow{5}{*}{\adjustbox{valign=c}{\rotatebox{90}{\textbf{IP2P}}}} 
        & A man wearing a denim jacket & make the jacket a rain coat & A man wearing a rain coat & \hl{make the coat a denim jacket} \\
        & A sofa in the living room & add pillows & A sofa in the living room with pillows & \hl{remove the pillows}  \\
        & $\cdots$ & $\cdots$ & $\cdots$ & $\cdots$  \\
        \midrule
        \multirow{4}{*}{\adjustbox{valign=c}{\rotatebox{90}{\textbf{CCXM}}}}
        & Person on the cover of a magazine & \hl{make the person a cat} & \hl{Cat on the cover of the magazine} & \hl{make the cat a person} \\
        & A tourist rests against a concrete wall & \hl{give him a backpack} & \hl{A tourist with a backpack rests against a concrete wall} & \hl{remove his backpack} \\
        & $\cdots$ & $\cdots$ & $\cdots$ & $\cdots$  \\
        \bottomrule
      \end{tabular}
  } %
\end{table*}

During training, we sample a \textit{noise level} $t$ from a predefined schedule and apply it to the input image, $I_i$, to obtained the noisy version. Based on the \textit{forward instruction}, $C_f$, \textit{Edit Model} then predicts and removes this noise to generate the \textit{edited image}, $I_e$. The \textit{reverse instruction}, $C_r$, is then applied, using another sampled noise level $\hat{t}$, to recover the original image, $I_r$. This \textit{single-step training approach} ensures that ERC is learned efficiently across varying noise scales, improving robustness across diverse edit scenarios.

By combining these three components—Text and Image Direction Alignment, Attention Map Alignment, and Reconstruction Alignment—our framework enforces ERC on real-image datasets \textit{without requiring triplets of input-edited images with edit instructions}. This enables our approach enables to scale effectively beyond synthetic datasets, making it a practical solution for real-world instruction-based image editing.

\subsubsection{Loss Functions}

To enforce ERC, we introduce loss terms that guide editing and reconstruction. Each training iteration processes a \textit{single-step training} sample consisting of an input image, an edit instruction, a reverse instruction, and captions.

\vspace{2pt}\noindent\textbf{CLIP Direction Loss.} 
This loss ensures that the transformation applied to the image aligns with the text instruction in CLIP's semantic space~\cite{Gal2022StyleGAN-NADA}. Given the CLIP embeddings of the input image ($E_{I_\text{i}}$), edited image ($E_{I_\text{e}}$), input caption ($E_{T_\text{i}}$), and edited caption ($E_{T_\text{e}}$), the loss is defined as:

\begin{equation}
\mathcal{L}_\text{CLIP} = 1 - \cos\left( E_{I_\text{e}} - E_{I_\text{i}}, E_{T_\text{e}} - E_{T_\text{i}} \right)
\end{equation}
By aligning the direction of change in image space with the transformation described in text space, this loss ensures that modifications accurately reflect the intended semantic edits.

\vspace{2pt}\noindent\textbf{Attention Map Alignment Loss.}  
To ensure that the same regions of the image are modified during both forward and reverse edits, we introduce an attention map alignment loss. Let $A_f^{(i)}$ and $A_r^{(i)}$ represent the cross-attention maps from the $i$-th layer of the U-Net model during the forward and reverse edits, respectively. The loss is defined as:

\begin{equation}
\mathcal{L}_\text{attn} = \sum_{i} \left\| A_f^{(i)} - A_r^{(i)} \right\|_2
\end{equation}
This loss ensures the model focuses on the same image regions during both forward and reverse edits—a key requirement for ERC. We compute it across U-Net’s down and up layers at random timesteps. By encouraging consistent cross-attention maps, the loss promotes spatial coherence across noise levels, as fixed instructions tend to yield stable attention over edited regions (see Supp. Sec. 4.2).

\vspace{2pt}\noindent\textbf{CLIP Similarity Loss.} 
This loss encourages the edited image to remain semantically aligned with the provided textual instruction. It is calculated as the cosine similarity between the CLIP embeddings of the edited image ($E_{I_\text{e}}$) and the edited caption ($E_{T_\text{e}}$):

\begin{equation}
\mathcal{L}_\text{sim} = 1 - \cos(E_{I_\text{e}}, E_{T_\text{e}})
\end{equation}
This loss ensures that the generated image aligns with the intended edits, maintaining semantic alignment with the provided instructions.

\vspace{2pt}\noindent\textbf{Reconstruction Loss.} To guarantee that the original image is recovered after the reverse edit, we employ a reconstruction loss consisting of a pixel-wise loss and a CLIP-based semantic loss. The total reconstruction loss is defined as:
\begin{equation}
\mathcal{L}_\text{recon} = \| I_\text{i} - I_\text{r} \|_2 + [1 - \cos(E_{I_\text{i}}, E_{I_\text{r}})]
\end{equation}
This loss ensures that the model can accurately reverse edits and return to the original image when the reverse instruction is applied, enforcing ERC by minimizing differences between the input and reconstructed images.

\vspace{2pt}\noindent\textbf{Total Loss.} Overall, all losses are applied to a \textit{single-step noise prediction}, making training more efficient. It is defined as a weighted combination of the individual losses:
\begin{equation}
\mathcal{L}_\text{ERC} = \lambda_\text{CLIP} \mathcal{L}_\text{CLIP} + \lambda_\text{attn} \mathcal{L}_\text{attn} + \lambda_\text{sim} \mathcal{L}_\text{sim} + \lambda_\text{recon} \mathcal{L}_\text{recon}
\end{equation}
where $\lambda_\text{CLIP}$, $\lambda_\text{attn}$, $\lambda_\text{sim}$, and $\lambda_\text{recon}$ are parameters that control the relative contributions of each loss term. The weights are selected through validation-based grid search to minimize $\mathcal{L}_{\text{ERC}}$ on a held-out set. Starting from the base configuration with $\mathcal{L}_{\text{CLIP}}$ and $\mathcal{L}_{\text{recon}}$, we found that $\mathcal{L}_{\text{sim}}$ enables freer edits by encouraging stronger image-text alignment, while $\mathcal{L}_{\text{attn}}$ improves localization by focusing on relevant regions during both forward and reverse edits.

\begin{figure*}[!htpb]
  \centering
  \includegraphics[width=\textwidth]{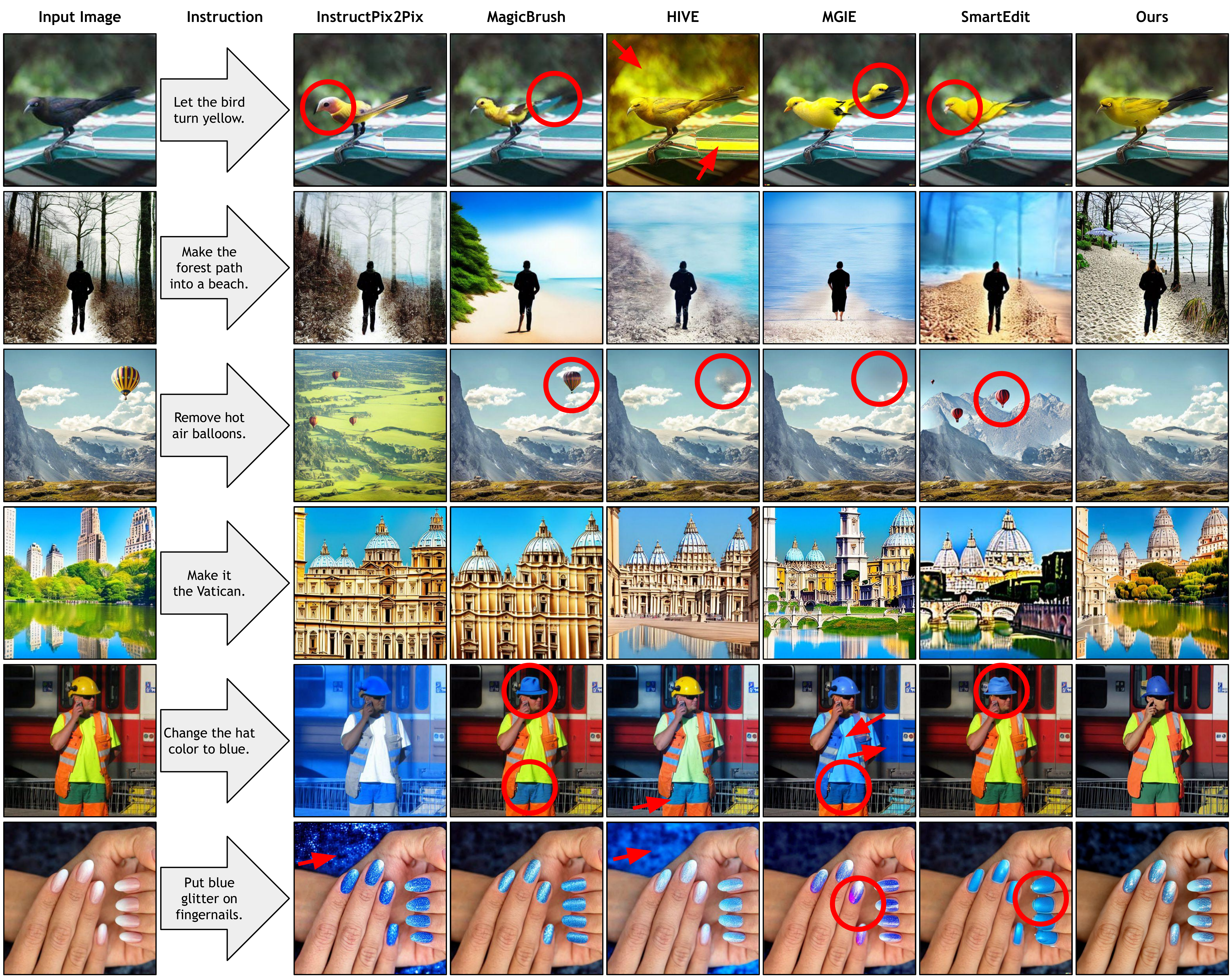}
\caption{\textbf{Qualitative Examples.} \model{} performance is shown across various tasks and datasets, compared to InstructPix2Pix, MagicBrush, HIVE, MGIE, and SmartEdit. Our method demonstrates either comparable or superior results in terms of accurately applying the requested edits while preserving visual consistency. Red circles and arrows indicate drastic problems during the image editing.}
  \label{fig:qualitatives}   
  \vspace{-0.75em}
\end{figure*}

\subsection{Training Data}
\label{subsec:data}
To enable ERC training on datasets with image and edit instructions~\cite{Brooks2022InstructPix2Pix}, we leverage Large Language Models (LLMs), such as GEMMA2~\cite{team2024gemma} and GEMINI~\cite{team2023gemini}, to automatically generate reverse edit instructions. These LLMs provide an efficient and scalable solution for obtaining reverse instructions with minimal cost and effort~\cite{Brooks2022InstructPix2Pix}. 
We use GEMINI to enrich the IP2P dataset with reverse instructions based on the input caption, edit instruction, and corresponding edited caption. To improve model performance, we employ few-shot prompting during this process, enabling the generation of reverse instructions without the need for manually paired datasets, which significantly enhances scalability. The reverse instructions generated by the LLM aim to revert the edited image to its original form (see~\cref{tab:samples} - IP2P section). 
Using the enriched dataset with reverse instructions  (see~\cref{tab:samples}, IP2P section), we fine-tune GEMMA2~\cite{team2024gemma}, to generate an edit instruction, edited caption, and reverse instruction given an input caption. We use this fine-tuned model to allow training on image-caption paired datasets such as CC3M and CC12M~\cite{sharma2018conceptual, changpinyo2021conceptual}, generating forward and reverse edits along with corresponding edited captions (see~\cref{tab:samples}, CCXM section).

\begin{figure*}
  \centering
  \hfill
  \begin{subfigure}{0.58\linewidth}
    \resizebox{\linewidth}{!}{
    \begin{tabular}{clccccc}
    \toprule
    \textbf{Settings}                     & \multicolumn{1}{c}{\textbf{Methods}} & L1$\downarrow$  & L2$\downarrow$  & CLIP-I$\uparrow$ & DINO$\uparrow$  & CLIP-T$\uparrow$ \\ \midrule
    \multirow{5}{*}{\textbf{Single-turn}} & HIVE~\cite{Zhang2023HIVE}    & 0.1092 & 0.0341 & 0.8519 & 0.7500 & 0.2752                              \\
      & InstructPix2Pix~\cite{Brooks2022InstructPix2Pix}    & 0.1122                             & 0.0371                             & 0.8524                               & 0.7428                             & 0.2764                               \\
      & \model{} w/ IP2P Dataset  & 0.0722 & 0.0193 & 0.9243 & 0.8876 & 0.2944                         \\ %
      & \model{} w/ HQ-Edit Dataset  & 0.0709 & 0.0190 & 0.9251 & 0.8882 & 0.2932                         \\ %
      & \model{} w/ CC3M Dataset  & \underline{0.0680} & \underline{0.0183} & \underline{0.9262} & \underline{0.8924} & \textbf{0.2966}                         \\ %
      & \model{} w/ CC12M Dataset & \textbf{0.0619} & \textbf{0.0174} & \textbf{0.9318} & \textbf{0.9039} & \underline{0.2964}                         \\ \midrule %
    \multirow{5}{*}{\textbf{Multi-turn}}  &  HIVE~\cite{Zhang2023HIVE}    & 0.1521 & 0.0557 & 0.8004 & 0.6463 & 0.2673                               \\
      & InstructPix2Pix~\cite{Brooks2022InstructPix2Pix}    & 0.1584                             & 0.0598                             & 0.7924                               & 0.6177                             & 0.2726                               \\
      & \model{} w/ IP2P Dataset   & 0.1104 & 0.0358 & 0.8779 & 0.8041 & 0.2892 \\ %
      & \model{} w/ HQ-Edit Dataset   & 0.1060 & 0.0344 & 0.8792 & 0.8081 & 0.2897 \\ %
      & \model{} w/ CC3M Dataset   & \underline{0.1040} & \underline{0.0337} & \underline{0.8816} & \underline{0.8130} & \textbf{0.2909} \\ %
      & \model{} w/ CC12M Dataset  & \textbf{0.0976} & \textbf{0.0323} & \textbf{0.8857} & \textbf{0.8235} & \underline{0.2901} \\ %
                                          \bottomrule
    \vspace{-0.5em}
    \end{tabular}    }
    \vspace{4pt}
        \caption{\textbf{Zero-shot Quantitative Comparison on MagicBrush~\cite{Zhang2023MagicBrush} test set.} Instruction-based editing methods that are not fine-tuned on MagicBrush are presented. In the multi-turn setting, target images are iteratively edited from the initial images. \label{tab:mb-quantitative}}
    \label{tab:quantitative-study}
  \end{subfigure}
  \hfill
  \begin{subfigure}{0.36\linewidth}
    \centering
    \includegraphics[width=\linewidth]{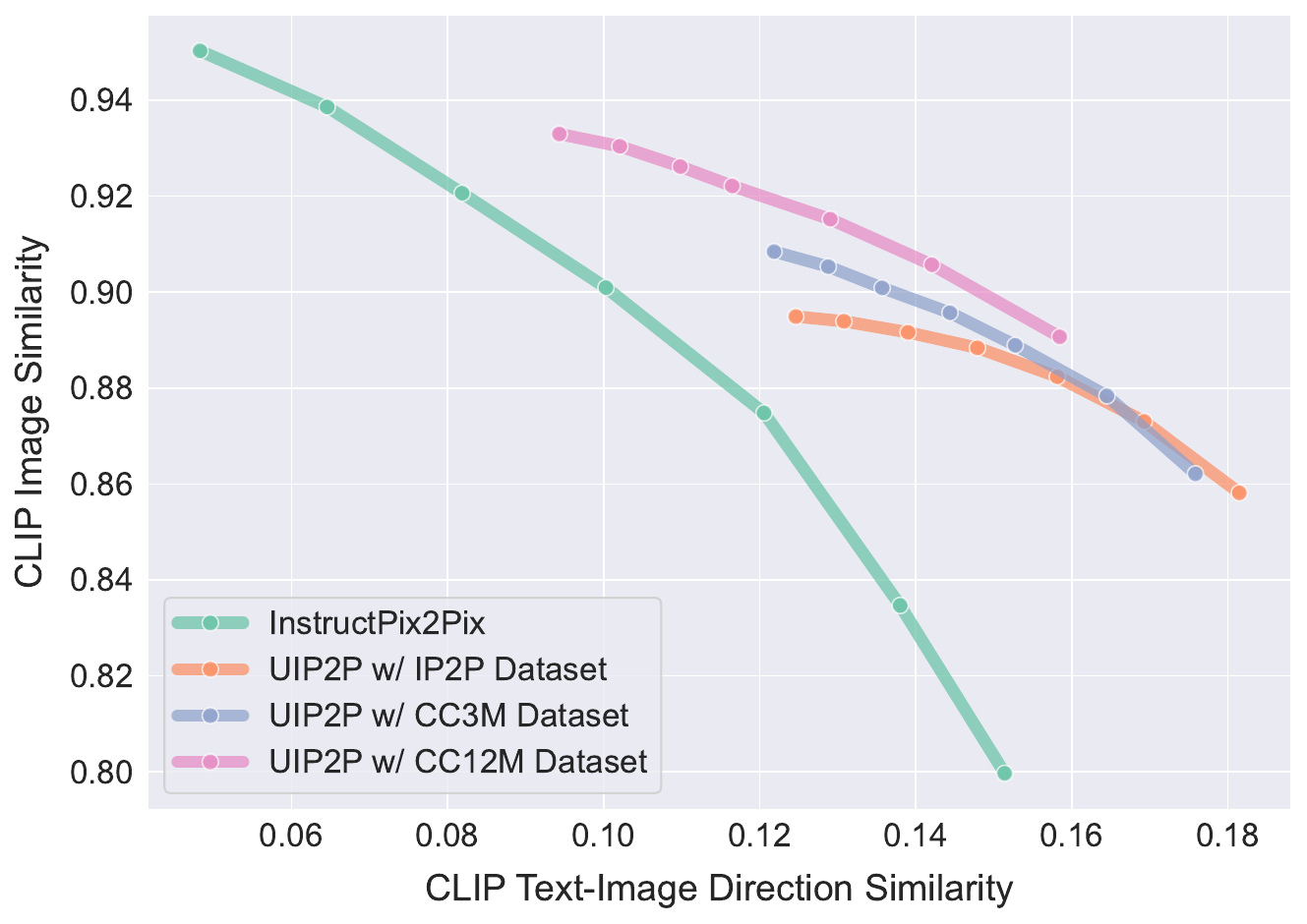}
    \caption{\textbf{Evaluation on the IP2P test dataset.} \model{} outperforms IP2P in both CLIP image similarity and CLIP text-image similarity metrics, demonstrating better visual fidelity and instruction alignment. } \label{fig:ip2p_eval}
  \end{subfigure}
  \vspace{-0.5em}
  \caption{Evaluation on MagicBrush and IP2P test datasets.} \vspace{-2em}
  \hfill
\end{figure*}

\section{Experiments}
\noindent\textbf{Datasets.}
We generate datasets consisting of forward and reverse instructions, as detailed in \cref{subsec:data}. For the initial experiments, we use the InstructPix2Pix dataset~\cite{Brooks2022InstructPix2Pix}, which provides generated image-caption pairs along with edit instructions. 
We further extend our experiments to real-image datasets, including CC3M~\cite{sharma2018conceptual} and CC12M~\cite{changpinyo2021conceptual}, for which we generate eight possible edits per image-caption pair. This increases diversity in the editing tasks, exposing the model to a wide range of transformations and enhancing its ability to generalize across different types of edits and real-world scenarios.
We then also conduct experiments on the HQ-Edit dataset~\cite{hui2025hqedit}, which already contains reverse instructions and higher-quality images, though they are still generated by off-the-shelf multimodal models \cite{dalle3, achiam2023gptv}.

\vspace{2pt}\noindent\textbf{Baselines.}
We evaluate our method by comparing it against several models. The primary baseline is InstructPix2Pix~\cite{Brooks2022InstructPix2Pix}, a supervised method that relies on ground-truth edited images during training. To demonstrate the advantages of our unsupervised approach, we train and test both IP2P and our model on the same datasets, with our method not using the ground-truth edited images while training. 
We also compare our method with other instruction-based editing models, including MagicBrush~\cite{Zhang2023MagicBrush}, HIVE~\cite{Zhang2023HIVE}, MGIE~\cite{fu2023guiding}, SmartEdit~\cite{huang2024smartedit}, and HQ-Edit~\cite{hui2025hqedit}. These comparisons allow us to evaluate how effectively our model handles diverse and complex edits without needing for editing methods to generate ground-truth edited images or human-annotated data.

\vspace{2pt}\noindent\textbf{Implementation Details.}
Our method, \model{}, fine-tunes the {SD-v1.5 checkpoint}~\cite{Rombach2022StableDiffusion} without any pre-training on triplet datasets. While we retain the {IP2P architecture} (\textit{Edit Model}, see \cref{fig:method}), our approach employs different training objectives, focusing primarily on enforcing the {ERC}. Specifically, we leverage the {CLIP ViT-L/14 model}, already integrated into SD-v1.5, to compute the losses.
Unlike supervised methods that begin training by adding noise to the ground-truth edited image, our method {starts by adding noise to the input image}. This design choice has significant implications, \eg{,} it helps bias the model toward preserving the structure and details of the original image, ensuring that edits remain consistent with the input. %
Our method trains in a {single denoising step}, predicting only one step per iteration to ensure efficiency. \model{} is trained using the {AdamW optimizer}~\cite{loshchilov2017decoupled} with a batch size of {256} over {11K iterations}. The base learning rate is set to {1e-05}. All experiments are conducted on {8 NVIDIA A100 GPUs}, with loss weights set as \(\lambda_\text{CLIP} = 1.0\), \(\lambda_\text{attn} = 0.5\), \(\lambda_\text{sim} = 1.0\), and \(\lambda_\text{recon} = 1.0\). The best configuration is selected based on the validation loss of $\mathcal{L}_{\text{ERC}}$.

\subsection{Qualitative Results}
\label{subsec:qual_results}

We compare \model{} with state-of-the-art methods, including InstructPix2Pix~\cite{Brooks2022InstructPix2Pix}, MagicBrush~\cite{Zhang2023MagicBrush}, HIVE~\cite{Zhang2023HIVE}, MGIE~\cite{fu2023guiding}, and SmartEdit~\cite{huang2024smartedit}, on various datasets~\cite{Brooks2022InstructPix2Pix, Zhang2023MagicBrush, Shi2020GIER, shi2021ma5k}. The tasks include color modifications, object removal, and structural changes. \model{} consistently produces high-quality edits, applying transformations accurately while maintaining visual coherence. For example, in ``let the bird turn yellow," \model{} provides a more natural color change while preserving the bird's shape. Similar improvements are observed in tasks like ``remove hot air balloons" and ``change hat color to blue." These results highlight \model{}’s ability to perform diverse edits, achieving successful and localized image modifications (see \cref{fig:qualitatives}).

\subsection{Quantitative Results}

\subsubsection{IP2P Test Dataset} 
We evaluate our method on the IP2P test split, containing 5K image-instruction pairs. Following \cite{Brooks2022InstructPix2Pix}, we use CLIP image similarity for visual fidelity and CLIP text-image similarity to assess alignment with the instructions. Higher scores in both metrics indicate better performance (upper right corner) by preserving image details (image similarity) and effectively applying the edits (direction similarity). As shown in \cref{fig:ip2p_eval}, \model{} outperforms IP2P across both metrics. In these experiments, the text scale $s_T$ is fixed, while the image scale $s_I$ varies from $1.0$ to $2.2$. 

\subsubsection{MagicBrush Test Dataset} The MagicBrush test split contains 535 sessions (source images for iterative editing) and 1053 turns (individual editing steps). It uses L1 and L2 norms for pixel accuracy, CLIP-I and DINO embeddings for image quality via cosine similarity, and CLIP-T to ensure alignment with local text descriptions. 
As seen in~\cref{tab:mb-quantitative}, \model{} performs the best for both single- and multi-turn settings. It is important to be noted that HIVE utilizes human feedback on edited images to understand user preferences and fine-tunes IP2P based on learned rewards, aligning the model more closely with human expectations.
\Cref{tab:mb-quantitative} shows that adding inverse instructions to IP2P improves performance, while training on HQ-Edit further enhances results due to its high-quality/resolution images and built-in reverse instructions. However, its smaller size ($\approx$100K) and specialized collection pipeline limit its impact. In contrast, leveraging large-scale real-image datasets like CC3M and CC12M enables scalable training without labor-intensive data collection, achieving the best overall performance. See Appendix for more quantitative comparison.

\subsubsection{User Study}
We conduct a user study on the Prolific platform~\cite{prolific} with 50 participants to evaluate six methods—IP2P, MagicBrush, HIVE, MGIE, SmartEdit, and \model{}—on 30 randomly sampled image-edit instructions from diverse datasets~\cite{Brooks2022InstructPix2Pix, Zhang2023MagicBrush, Shi2020GIER, shi2021ma5k}.
Participants compare images from randomly paired methods, selecting the better-performing one across six pairs per question, yielding \textit{18,000} total votes—9,000 for \textit{edit success} and 9,000 for \textit{localization accuracy}.
In two runs, they first choose the model best following the instruction, \textit{Successful}, then the one applying edits only to the intended area, \textit{Localized}. To ensure robust evaluation, we use \textit{Elo rating system} \cite{elo1978rating}, a widely adopted ranking method in LLM and generative model benchmarking \cite{chatbot_arena, elo_uncovered}. By enabling continuous pairwise comparisons, Elo accounts for varying instruction difficulty and produces a relative ranking. \Cref{tab:user_study} shows that \model{} achieves the highest \textit{Elo score} and adjusted \textit{Win rate} across both criteria. See Appendix for more details.

\begin{table}[!htb]
    \centering     
    \captionof{table}{\textbf{User study.} Results comparing six methods based on Elo scores and top performer ratios for edit success and localization accuracy. \model{} achieves the highest scores in both criteria.} 
    \vspace{-0.5em}
    \label{tab:user_study}
    \resizebox{0.9\linewidth}{!}{
    \begin{tabular}{c|cc|cc}
    \toprule 
    & \multicolumn{2}{c|}{Successful}& \multicolumn{2}{c}{Localized}\\
    \midrule
    Models & Elo score & Win rate & Elo score & Win rate \\
    \midrule
    IP2P & 1395 & 12\% & 1301 & 12\% \\
    MagicBrush & 1482 & 16\% & \underline{1601} & \underline{18\%} \\
    HIVE & 1465 & 15\% & 1471 & 14\% \\
    MGIE & \underline{1584} & \underline{18\%} & 1529 & 17\% \\
    SmartEdit & 1543 & 17\% & 1440 & 16\% \\
    \model{} (Ours) & \textbf{1681} & \textbf{22\%} & \textbf{1659} & \textbf{23\%} \\ 
    \bottomrule
    \end{tabular}
    }
\end{table}

\subsection{Ablation Study} \label{subsec:ablation}
\noindent\textbf{Scalability.} We train our method on IP2P, HQ-Edit, CC3M, and CC12M to assess its scalability with dataset size and quality. As shown in \cref{tab:quantitative-study}, performance improves with larger and higher-quality datasets, with notable gains from IP2P to CC12M. This highlights our method’s ability to leverage extensive data for better fidelity and localization. See Appendix for the discussion on runtime of our method.

\vspace{2pt}\noindent\textbf{Loss functions.}
We conduct a zero-shot evaluation on the MagicBrush test set (single-turn) to assess the contribution of each loss function, see \cref{table:loss} - Loss. Starting with the base configuration, containing $\mathcal{L}_{CLIP}$ and $\mathcal{L}_{recon}$, we observe moderate performance across the same metrics. Adding $\mathcal{L}_{sim}$ loss allows the model to perform edits more freely, as the \textit{Base} without it tends to create outputs similar to the input image. Finally, $\mathcal{L}_{attn}$ enhances the model's focus on relevant regions and ensures that the region will remain consistent between the forward and reverse processes.

\begin{table}[!ht]
\centering 
    \caption{\textbf{Ablation studies.} Adding additional loss functions to the base setup improves performance on the MagicBrush benchmark. Additionally, using random timestamps between forward and reverse instructions leads to better edit localization and overall performance.} \label{table:loss}    
    \vspace{-0.5em}

    \resizebox{\linewidth}{!}{
        \begin{tabular}{c|lccccc}
        \toprule
        & \textbf{Ablation} & L1$\downarrow$  & L2$\downarrow$  & CLIP-I$\uparrow$ & DINO$\uparrow$  & CLIP-T$\uparrow$ \\ \midrule
            & IP2P~\cite{Brooks2022InstructPix2Pix} & 0.112 & 0.037 & 0.852 & 0.743 & 0.276 \\ \midrule
            \multirow{3}{*}{\rotatebox[origin=c]{90}{%
            \begin{tabular}{@{}c@{}}
                \textbf{Loss}
            \end{tabular}%
        }} & Base & 0.117 & 0.032 & 0.878 & 0.806 & \textbf{0.309} \\
            & + $\mathcal{L}_{sim}$  & 0.089 & 0.024 & 0.906 & 0.872 & 0.301 \\
            & + $\mathcal{L}_{attn}$ & \textbf{0.062} & \textbf{0.017} & \textbf{0.932} & \textbf{0.904} & {0.296}                         \\ \midrule
        \multirow{2}{*}{\rotatebox[origin=c]{90}{%
            \begin{tabular}{@{}c@{}}
                \textbf{Attn}
            \end{tabular}%
        }} & Same $t$ & 0.074 & 0.022 & 0.921 & 0.889 & \textbf{0.299} \\
            & Random $t$ & \textbf{0.062} & \textbf{0.017} & \textbf{0.932} & \textbf{0.904} & {0.296} \\
            \bottomrule
        \end{tabular}
    }     

\end{table}

\vspace{2pt}\noindent\textbf{Attention Map Alignment Loss vs. Timestamp $t$.}
We compare the impact of using random timestamps $t$ versus same timestamps $t$ between forward and reverse instructions. As shown in \cref{table:loss} - Attn, using random $t$ consistently outperforms the fixed approach across all metrics. This suggests that introducing variability in $t$ during training leads to a more consistent cross-attention, resulting in better edit localization and overall performance. The rationale is that using different noise levels enhances robustness by exposing the model to varied denoising conditions during forward and reverse edits, while the attention consistency loss ensures the model maintains focus on the same regions regardless of the noise level. As also investigated in \cite{guo2024focus, simsar2023lime}, maintaining consistent attention maps across different timestamps enables more precise editing.

\section{Conclusion}

In this work, we introduce \model{}, a novel unsupervised approach for instruction-based image editing that eliminates the need for paired supervision through the ERC. By enabling training on real-image datasets, our approach enhances scalability while maintaining high-fidelity edits. Through a single-step training paradigm, we achieve efficient and precise edits, outperforming existing methods.
While our method advances instruction-based editing, future improvements in reverse instruction generation, spatial guidance, and diffusion model efficiency can further enhance performance. Additionally, incorporating perceptual or adversarial losses could improve photorealism, addressing cases where optimizing for CLIP alignment may compromise visual quality in challenging edits. As models continue to evolve, our approach stands to benefit from these advancements, further improving edit precision and inference speed.
Overall, our work unlocks new possibilities for large-scale training in instruction-based image editing without explicit supervision. We encourage future research to build upon this foundation, refining both model design and training strategies.

{
    \small
    \bibliographystyle{ieeenat_fullname}
    \bibliography{main}
}

\newpage
\appendix
\section*{Table of Contents}
\startcontents[appendices]
\setcounter{tocdepth}{2} %
\printcontents[appendices]{l}{1}{\setcounter{tocdepth}{2}}

\section{Ethics Statement}
Advancements in localized image editing technology offer substantial opportunities to enhance creative expression and improve accessibility within digital media and virtual reality environments. Nonetheless, these developments also bring forth important ethical challenges, particularly concerning the misuse of such technology to create misleading content, such as deepfakes~\cite{korshunov2018deepfakes}, and its potential effect on employment in the image editing industry. Moreover, as also highlighted by \cite{kenthapadi2023generative}, it requires a thorough and careful discussion about their ethical use to avoid possible misuse. We believe that our method could help reduce some of the biases present in previous datasets, though it will still be affected by biases inherent in models such as CLIP. Ethical frameworks should prioritize encouraging responsible usage, developing clear guidelines to prevent misuse, and promoting fairness and transparency, particularly in sensitive contexts like journalism. Effectively addressing these concerns is crucial to amplifying the positive benefits of the technology while minimizing associated risks. In addition, our user study follows strict anonymity rules to protect the privacy of participants.

\section{Runtime Analysis}

Our method modifies the training objectives of IP2P by incorporating Edit Reversibility Constraint (ERC) and additional loss functions. However, these changes do not affect the overall runtime. Inference time remains comparable to the original IP2P framework, as we retain the same architecture and model structure. Consequently, our approach introduces no additional complexity or overhead in terms of processing time or resource consumption. This gives \model{} an advantage over methods like MGIE~\cite{fu2023guiding} and SmartEdit~\cite{huang2024smartedit}, which rely on large language models (LLMs) during inference in terms of runtime and resource consumption.

\section{Elo Rating System}
The Elo rating system is a widely used method for ranking competitors in pairwise comparisons, originally designed for chess and later adopted in various domains, including generative model evaluation. It assigns each candidate a score that dynamically updates based on comparative performance. A higher Elo score indicates stronger performance relative to other candidates.

\subsection{Implementation Details}
Our implementation follows the standard Elo rating mechanism with the following key components:
\begin{itemize}
    \item \textbf{Initial Rating:} Each model starts with a rating of \textbf{1500}.
    \item \textbf{K-Factor ($K = 32$):} Governs the magnitude of rating updates.
    \item \textbf{Expected Score Calculation:} The probability of a model winning against another is computed as:
    \begin{equation}
        E_A = \frac{1}{1 + 10^{(R_B - R_A)/400}}
    \end{equation}
    where $R_A$ and $R_B$ are the Elo ratings of the two competing models.
    \item \textbf{Score Update Rule:} After each comparison, the winner’s rating increases, and the loser’s rating decreases:
    \begin{equation}
        R_A' = R_A + K (S_A - E_A)
    \end{equation}
    where $S_A = 1$ for a win, $0$ for a loss, and $0.5$ for a draw.
\end{itemize}

\subsection{Interpreting Elo Scores}
Elo scores provide an intuitive understanding of model performance:
\begin{itemize}
    \item \textbf{Higher Scores:} Indicate models that consistently outperform others.
    \item \textbf{Score Differences:} A gap of 400 points implies the higher-rated model is \textbf{10 times more likely} to win.
    \item \textbf{Stability of Rankings:} As the number of votes increases, the Elo system converges, producing a reliable performance ranking.
\end{itemize}

By leveraging the Elo rating system, we ensure a \textbf{robust, adaptive, and comparative} evaluation framework for image-editing models. This approach provides a dynamic ranking that accounts for dataset variations and human annotation biases, aligning with best practices in benchmarking generative models.

\section{Additional Details on Ablation Studies}

\subsection{Loss Functions}

\begin{figure}[!htb]
    \centering
    \includegraphics[width=0.85\linewidth]{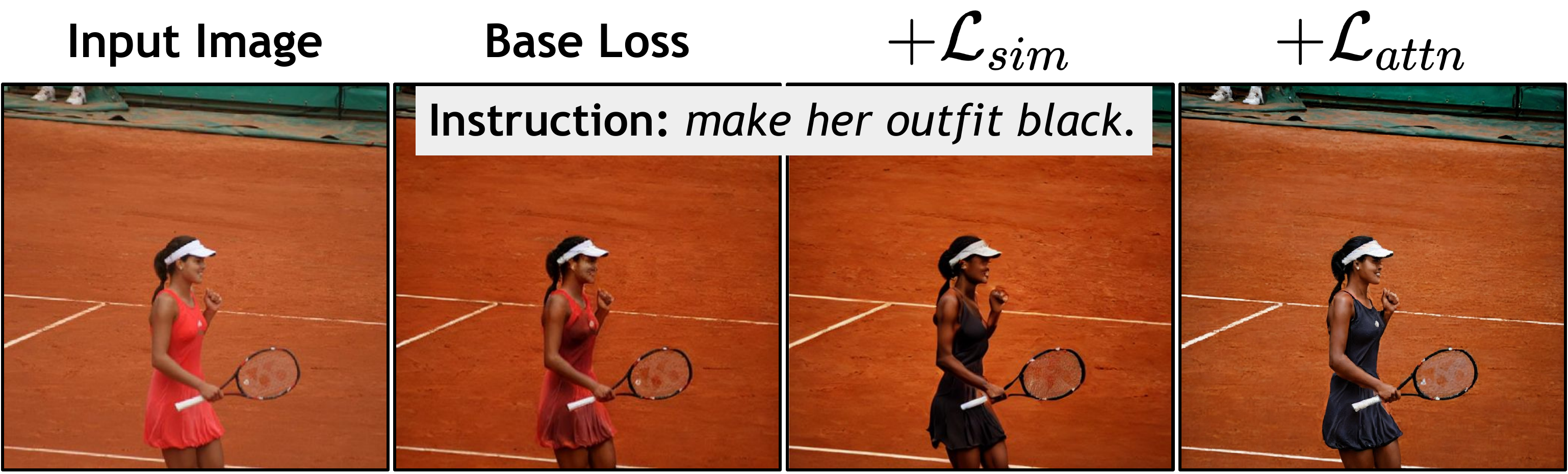}
    \caption{\textbf{Visual effects of loss components.} This figure demonstrates how different loss components affect the final editing results. The comparison shows the impact of various loss terms on edit quality and localization.}
    \label{fig:loss_components}
\end{figure}

We focused our ablation studies on $ \mathcal{L}_{\text{sim}} $ and $ \mathcal{L}_{\text{attn}} $ because these losses are additional components beyond the core $ \mathcal{L}_{\text{CLIP}} $ and $ \mathcal{L}_{\text{recon}} $. The core losses are essential for ensuring semantic alignment and reversibility in Edit Reversibility Constraint (ERC), forming the foundation of our method. Without $ \mathcal{L}_{\text{CLIP}} $ and $ \mathcal{L}_{\text{recon}} $, the model risks diverging, losing its ability to preserve both the input's structure and its semantic coherence during edits.

Adding $ \mathcal{L}_{\text{sim}} $ enables the model to perform edits more freely by encouraging alignment between image and textual embeddings, thereby expanding its capacity for complex and diverse transformations. On the other hand, $ \mathcal{L}_{\text{attn}} $ refines the model's ability to focus on relevant regions during edits, improving localization and reducing unintended changes in non-targeted areas.

$\mathcal{L}_{\text{CLIP}} $ is applied between the input image and the edited image to ensure semantic alignment with the edit instruction. The reconstructed image is already constrained by $ \mathcal{L}_{\text{recon}} $, which enforces structural and semantic consistency with the input. Adding $ \mathcal{L}_{\text{CLIP}} $ to the reconstructed image would be redundant and could interfere with the reversibility objective. Our design does not apply $ \mathcal{L}_{\text{CLIP}} $ to the reconstructed image to preserve the focus on reversibility and prevent conflicting optimization objectives.

\subsection{Attention Across Noise Steps in Training}

\begin{figure}[!htb]
    \centering
    \includegraphics[width=\linewidth]{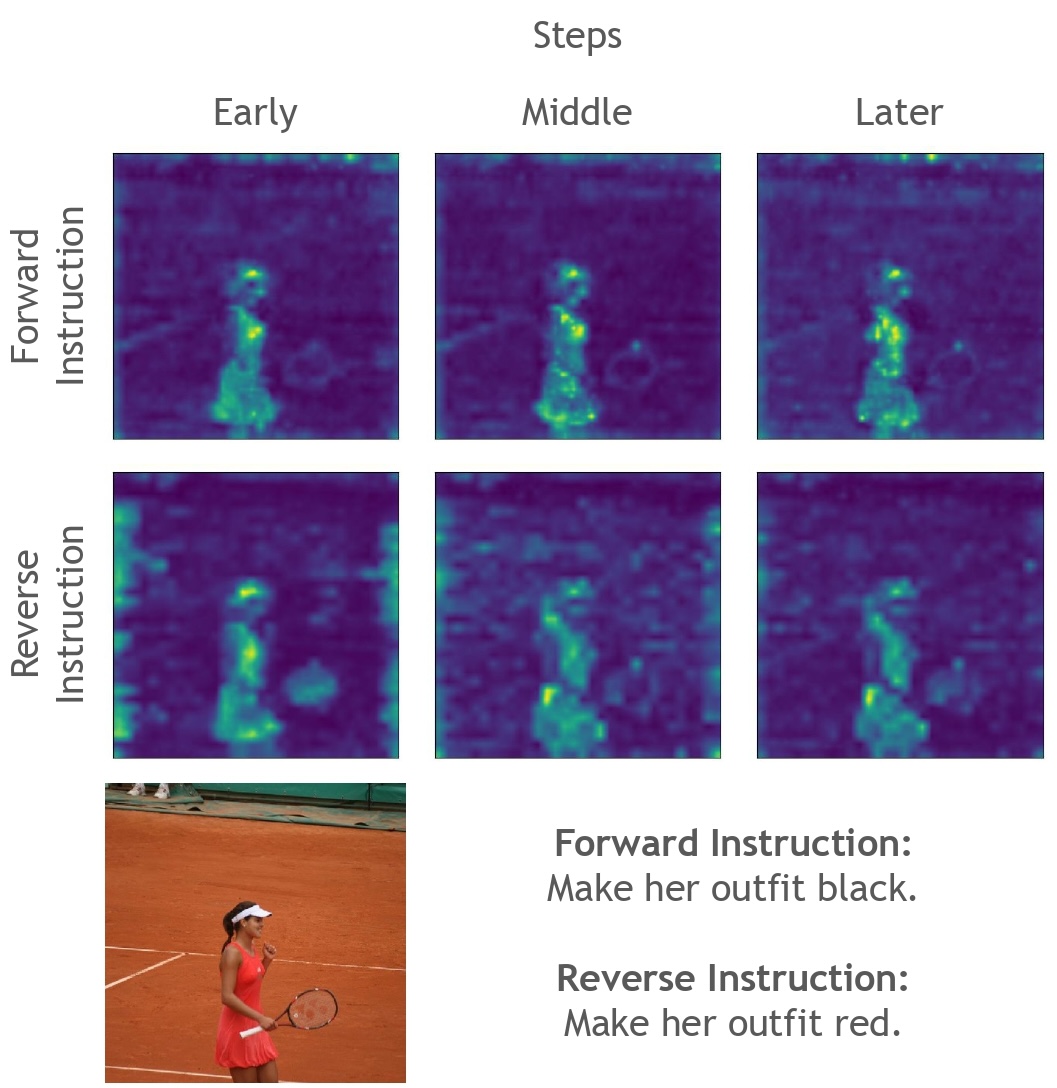}
    \caption{{\textbf{Attention maps for diffusion steps.} Cross-attention maps for forward (top) and reverse (middle) instructions across early, middle, and later noise steps. The model enforces attention consistency, focusing on relevant regions for both edits.}}    \label{fig:attn}
\end{figure}

{At training time, we sample two different noise steps for the forward and backward processes, which are conditioned on the input image and edit instruction. Attention consistency is enforced between these different noise steps to ensure that the model attends to the same regions during both forward and reverse edits. This is supported by the observation that cross-attention scores in instruction-based editing methods tend to be more consistent across timesteps, as the edit instruction remains fixed and the model's focus shifts only to the regions being edited (see Fig.} \ref{fig:attn}{)}.

{Recent works, such as those by Guo et al.}~\cite{guo2024focus} {and Simsar et al.}~\cite{simsar2023lime}{, demonstrate that regularizing attention space with a mask during inference enables localized edits in IP2P. Our method builds on these ideas by incorporating attention consistency into the training phase, making it possible to focus on relevant regions from the start and avoiding the need for additional inference-time modifications.}

\begin{figure*}[!htb]
    \centering
    \includegraphics[width=\textwidth]{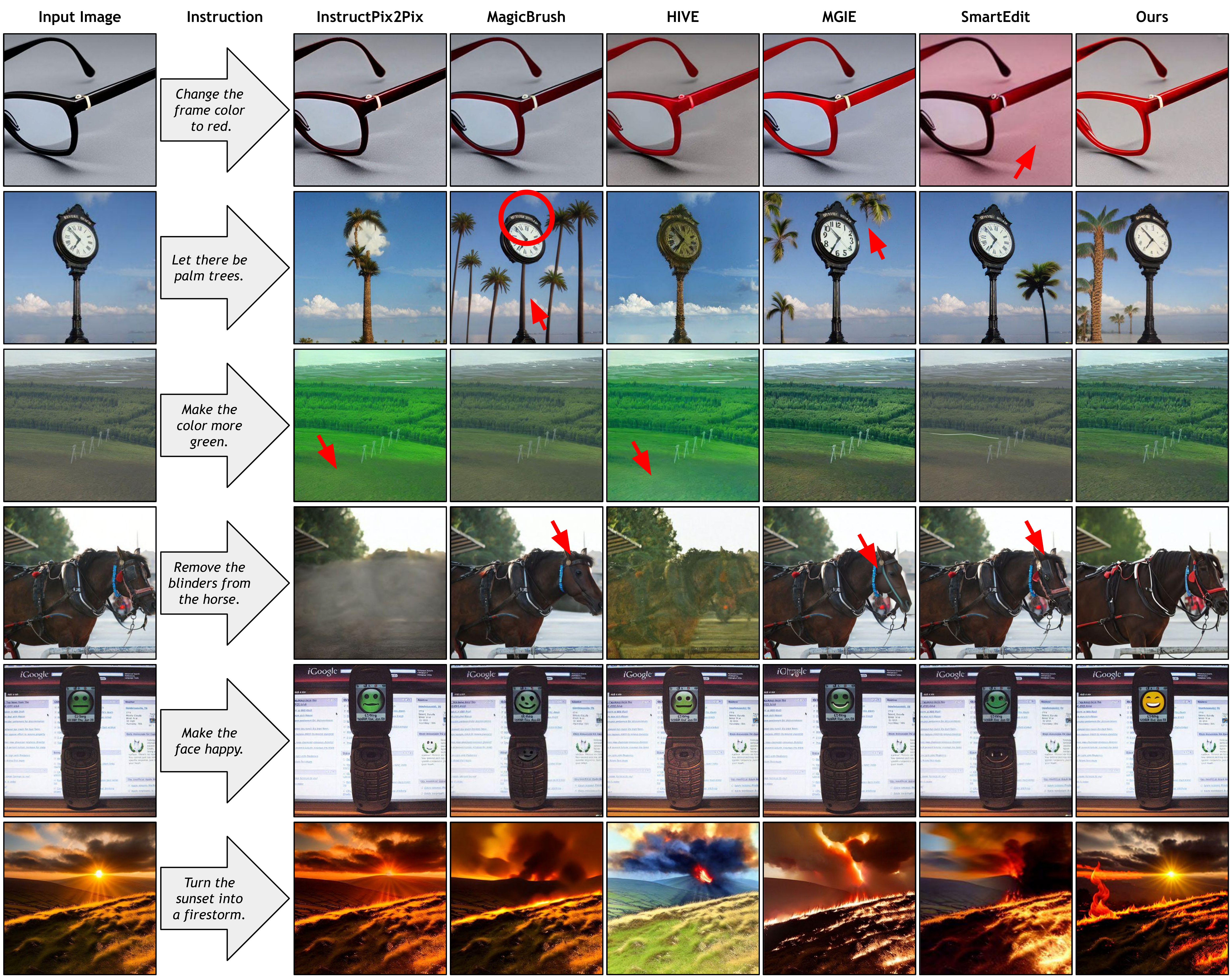}
    \caption{\textbf{Qualitative comparison of our method with baseline models for various editing instructions.} From left to right: Input image, edit instruction, and results from InstructPix2Pix, MagicBrush, HIVE, MGIE, SmartEdit, and our method. Our approach demonstrates superior fidelity and alignment with the provided instructions across diverse tasks, such as expression changes, color adjustments, object transformations, and creative edits. Red circles and arrows indicate drastic problems during the image editing.}
    \label{fig:qual2}
\end{figure*}

\section{Additional Qualitative Results}

To further demonstrate the capabilities of our approach, we present additional qualitative comparisons in ~\cref{fig:qual2}. These results showcase the performance of our method against several baseline models, including InstructPix2Pix, MagicBrush, HIVE, MGIE, and SmartEdit, across a diverse set of editing instructions. These tasks range from simple edits, such as color adjustments and expression changes, to more challenging transformations, including object removal, style changes, and complex scene edits.

The comparison highlights that our method consistently achieves higher fidelity and better alignment with the provided instructions. For instance, when instructed to modify facial expressions, such as ``make the face happy," our method produces more natural and expressive results. Similarly, for color adjustments, such as ``make the color more green," our approach ensures vibrant and accurate edits that surpass the performance of baseline models. In more challenging scenarios, like ``turn the sunset into a firestorm," our method maintains the structural integrity of the original image while executing the desired transformations. 
Furthermore, in creative edits, such as ``remove the blinders from the horse," our model demonstrates exceptional precision and attention to detail.

\section{Additional Quantitative Results}
\subsection{Evaluation on PIE-Bench}
We apply our method to the PIE benchmark \cite{ju2023direct} to evaluate its performance on diverse editing tasks and compare it to IP2P, a representative feed-forward instruction-based editing method and a supervised alternative to our approach. \Cref{tab:pie_comparison} summarizes the results.
The results show that our method outperforms IP2P across most metrics, including better preservation of structure (PSNR and SSIM), lower perceptual differences (LPIPS), and reduced mean squared error (MSE). These improvements demonstrate the scalability and versatility of our approach on a broader benchmark. This analysis is included in the revised manuscript to provide a comprehensive evaluation of our method.

\begin{table}[!htb]
    \centering
    \caption{{\textbf{Performance comparison on the PIE benchmark.} Lower values for Distance, LPIPS, and MSE indicate better performance, while higher values for PSNR, SSIM, Whole, and Edit indicate improved quality and structural preservation.}} %

    \resizebox{\linewidth}{!}{
    \begin{tabular}{lccccccc}
        \toprule
        \textbf{Methods} & \textbf{Distance ↓} & \textbf{PSNR ↑} & \textbf{LPIPS ↓} & \textbf{MSE ↓} & \textbf{SSIM ↑} & \textbf{Whole ↑} & \textbf{Edit ↑} \\
        \midrule
        {InstructDiffusion} & 75.44 & 20.28 & 155.66 & 349.66 & 75.53 & 23.26 & 21.34 \\
        {IP2P} & 57.91 & 20.82 & 158.63 & 227.78 & 76.26 & 23.61 & 21.64 \\
        {Ours} & \textbf{27.05} & \textbf{26.85} & \textbf{60.57} & \textbf{40.07} & \textbf{83.69} & \textbf{24.78} & \textbf{21.89} \\
        \bottomrule
    \end{tabular}
    }
    \label{tab:pie_comparison}
\end{table}

\subsection{Evaluation on Emu Edit}
We apply our method to the Emu Edit benchmark \cite{sheynin2023emu} to evaluate its performance on diverse editing tasks and compare it to other baselines. \Cref{tab:emu_comparison} presents a quantitative comparison across multiple metrics, assessing the quality of the generated edits. 
Our approach achieves the best results across most metrics, including higher CLIP scores, which indicate better alignment with textual descriptions, and improved perceptual quality as measured by DINO. Notably, our method surpasses all baselines in CLIP$_{dir}$ and CLIP$_{im}$, demonstrating stronger instruction adherence and output realism. The improvements in these objective measures further support the effectiveness of our method in complex editing tasks.

\begin{table}[ht!]
\centering
\caption{\textbf{Performance comparison on the Emu Edit benchmark.}Comparison with image-editing baselines evaluated on Emu Edit test set. For each method, we report CLIP, L1, and DINO metrics.} 
\label{tab:emu_comparison}
\resizebox{\linewidth}{!}{
\centering
\begin{tabular}{lccccc}
\toprule
Method & $\text{CLIP}_{dir}\!\uparrow$ &  $\text{CLIP}_{im}\!\uparrow$ & $\text{CLIP}_{out}\!\uparrow$ &  $\text{L1}\!\downarrow$  & DINO$\uparrow$ \\
\midrule
InstructPix2Pix~\cite{Brooks2022InstructPix2Pix} & 0.078 & 0.834 & 0.219 & 0.121 & 0.762 \\
MagicBrush~\cite{Zhang2023MagicBrush}      & 0.090 & 0.838 & 0.222 & 0.100 & 0.776 \\
PnP~\cite{tumanyan2023plug}             & 0.028 & 0.521 & 0.089 & 0.304 & 0.153 \\
Null-Text Inv.~\cite{Mokady2022NullTextInversion} & 0.101 & 0.761 & {0.236} & \textbf{0.075} & 0.678 \\
Emu Edit \cite{sheynin2023emu}             & {0.109} & {0.859} & 0.231 & 0.094 & {0.819} \\
Ours & \textbf{0.115} & \textbf{0.867} & \textbf{0.244} & 0.083 & \textbf{0.834} \\
\bottomrule
\end{tabular}}

\end{table}

\subsection{Evaluation on MagicBrush Test}
In this section, we present the full quantitative analysis on the MagicBrush test set, including results from both global description-guided and instruction-guided models, as shown in \cref{tab:mball-quantitative}. While our method, \model{}, is not fine-tuned on human-annotated datasets like MagicBrush, it still achieves highly competitive results compared to models specifically fine-tuned for the task. In particular, \model{} demonstrates either the best or second-best performance in key metrics such as L1, L2, and CLIP-I, even outperforming fine-tuned models in several cases. This highlights the robustness and generalization capabilities of \model{}, showing that it can effectively handle complex edits without the need for specialized training on real datasets. These results further validate that \model{} delivers high-quality edits in a variety of contexts, maintaining competitive performance against fine-tuned models on the MagicBrush dataset, which is human-annotated.

\begin{table}[thb!]
\small
\centering
\caption{\textbf{Quantitative comparison on MagicBrush~\cite{Zhang2023MagicBrush} test set.} In the multi-turn setting, target images are iteratively edited from the initial source images. 
Best results are in \textbf{bold}.}
\resizebox{\linewidth}{!}{
\begin{tabular}{clccccc}
\toprule
\textbf{Settings}                     & \multicolumn{1}{c}{\textbf{Methods}} & L1$\downarrow$  & L2$\downarrow$  & CLIP-I$\uparrow$ & DINO$\uparrow$  & CLIP-T$\uparrow$ \\ \midrule
\multirow{15}{*}{\textbf{Single-turn}} & \multicolumn{6}{c}{\textit{\textbf{Global Description-guided}}}                                                                                                                                                   \\ \cmidrule{2-7} 
  & Open-Edit~\cite{Liu2020Open-Edit}           & 0.1430                             & 0.0431                             & 0.8381                               & 0.7632                             & 0.2610                               \\
  & VQGAN-CLIP~\cite{Crowson2022VQGAN-CLIP}          & 0.2200                             & 0.0833                             & 0.6751                               & 0.4946                             & \textbf{0.3879}                      \\
  & SD-SDEdit~\cite{Meng2022SDEdit}           & 0.1014                             & 0.0278                             & 0.8526                               & 0.7726                             & 0.2777                               \\
  & Text2LIVE~\cite{Bar-Tal2022Text2LIVE}           & 0.0636                    & \textbf{0.0169}                    & 0.9244                         & 0.8807                      & 0.2424                               \\
  & Null Text Inversion~\cite{Mokady2022NullTextInversion} & 0.0749                             & 0.0197                       & 0.8827                               & 0.8206                             & 0.2737                               \\ \cmidrule{2-7} 
  & \multicolumn{6}{c}{\textit{\textbf{Instruction-guided}}}                                                                                                                                                          \\ \cmidrule{2-7} 
  & HIVE~\cite{Zhang2023HIVE}    & 0.1092 & 0.0341 & 0.8519 & 0.7500 & 0.2752                              \\
  & ~~~~w/ MagicBrush~\cite{Zhang2023MagicBrush}   & 0.0658 & 0.0224 & 0.9189 & 0.8655 & 0.2812                         \\ 
  & InstructPix2Pix~\cite{Brooks2022InstructPix2Pix}    & 0.1122                             & 0.0371                             & 0.8524                               & 0.7428                             & 0.2764                               \\
  & ~~~~w/ MagicBrush~\cite{Zhang2023MagicBrush}   & \underline{0.0625}                       & 0.0203                       & \textbf{0.9332}                      & \underline{0.8987}                    & 0.2781                         \\ 
  & \model{} w/ IP2P Dataset  & 0.0722 & 0.0193 & 0.9243 & 0.8876 & 0.2944                         \\ %
  & \model{} w/ CC3M Dataset  & 0.0680 & 0.0183 & 0.9262 & 0.8924 & \underline{0.2966}                         \\ %
  & \model{} w/ CC12M Dataset & \textbf{0.0619} & \underline{0.0174} & \underline{0.9318} & \textbf{0.9039} & 0.2964                         \\ \midrule %
\multirow{15}{*}{\textbf{Multi-turn}}  & \multicolumn{6}{c}{\textit{\textbf{Global Description-guided}}}                                                                                                                                                   \\ \cmidrule{2-7} 
  & Open-Edit~\cite{Liu2020Open-Edit}           & 0.1655                             & 0.0550                             & 0.8038                               & 0.6835                             & 0.2527                               \\
  & VQGAN-CLIP~\cite{Crowson2022VQGAN-CLIP}          & 0.2471                             & 0.1025                             & 0.6606                               & 0.4592                             & \textbf{0.3845}                      \\
  & SD-SDEdit~\cite{Meng2022SDEdit}           & 0.1616                             & 0.0602                             & 0.7933                               & 0.6212                             & 0.2694                               \\
  & Text2LIVE~\cite{Bar-Tal2022Text2LIVE}           & 0.0989                       & \textbf{0.0284}                    & 0.8795                         & 0.7926                       & 0.2716                               \\
  & Null Text Inversion~\cite{Mokady2022NullTextInversion} & 0.1057                             & 0.0335                       & 0.8468                               & 0.7529                             & 0.2710                               \\ \cmidrule{2-7} 
  & \multicolumn{6}{c}{\textit{\textbf{Instruction-guided}}}                                                                                                                                                          \\ \cmidrule{2-7} 
  & HIVE~\cite{Zhang2023HIVE}    & 0.1521 & 0.0557 & 0.8004 & 0.6463 & 0.2673                               \\
  & ~~~~w/ MagicBrush~\cite{Zhang2023MagicBrush}   & \underline{0.0966} & 0.0365 & 0.8785 & 0.7891 & 0.2796                       \\  
  & InstructPix2Pix~\cite{Brooks2022InstructPix2Pix}    & 0.1584                             & 0.0598                             & 0.7924                               & 0.6177                             & 0.2726                               \\
  & ~~~~w/ MagicBrush~\cite{Zhang2023MagicBrush}   & \textbf{0.0964}                    & 0.0353                             & \textbf{0.8924}                      & \textbf{0.8273}                    & 0.2754                         \\ 
  & \model{} w/ IP2P Dataset   & 0.1104 & 0.0358 & 0.8779 & 0.8041 & 0.2892 \\ %
  & \model{} w/ CC3M Dataset   & 0.1040 & 0.0337 & 0.8816 & 0.8130 & \underline{0.2909} \\ %
  & \model{} w/ CC12M Dataset  & {0.0976} & \underline{0.0323} & \underline{0.8857} & \underline{0.8235} & 0.2901 \\ %
                                      \bottomrule
\label{tab:mball-quantitative}
\end{tabular}
}
\end{table}

\begin{table*}[!htb]
  \caption{\textbf{Examples of Four Possible Edits for Two Different Input Captions.} Our dataset generation process showcases the flexibility of the reverse instruction dataset by demonstrating multiple transformations for the same caption.}
  \centering
  \resizebox{\textwidth}{!}{
    \begin{tabular}{@{}p{0.3\linewidth}@{\hskip 0.05in}>{\raggedright}p{0.2\linewidth}@{\hskip 0.05in}>{\raggedright}p{0.25\linewidth}@{\hskip 0.05in}>{\raggedright\arraybackslash}p{0.2\linewidth}@{}}
        \toprule
         Input Caption & Edit Instruction & Edited Caption & Reverse Instruction\\
        \midrule
        \multirow{7}{*}{A dog sitting on a couch} & change the dog’s color to brown & A brown dog sitting on a couch & change the dog’s color back to white \\
        & add a ball next to the dog & A dog sitting on a couch with a ball & remove the ball \\
        & remove the dog & An empty couch & add the dog back \\
        & move the dog to the floor & A dog sitting on the floor & move the dog back to the couch \\
        \midrule
        \multirow{7}{*}{A car parked on the street} & change the car color to red & A red car parked on the street & change the car color back to black \\
        & add a bicycle next to the car & A car parked on the street with a bicycle & remove the bicycle \\
        & remove the car & An empty street & add the car back \\
        & move the car to the garage & A car parked in the garage & move the car back to the street \\
        \bottomrule
    \end{tabular}
  }
\end{table*}

\section{Edit Reversibility Constraint Example}

\begin{figure}[!htb]
    \centering
    \includegraphics[width=\linewidth]{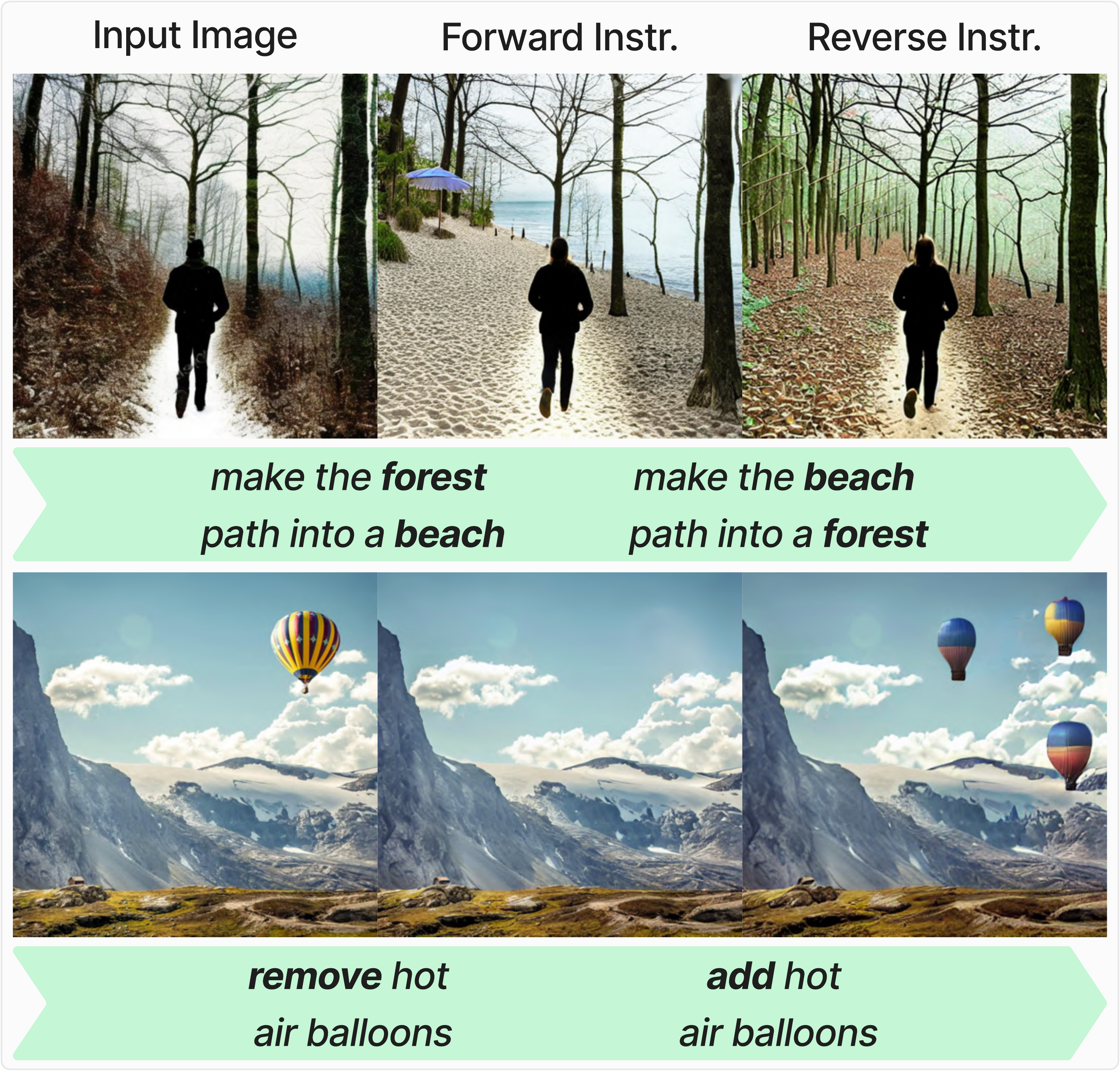}
    \caption{Forward and reverse edits are applied sequentially.}\label{fig:reverse_exp}
\end{figure}

We demonstrate ERC with a visual example during inference. In the forward pass, the model transforms the input image based on the instruction (\eg{,} ``turn the forest path into a beach"). In the reverse pass, the corresponding reverse instruction (\eg{,} ``turn the beach back into a forest") is applied, reconstructing the original image. This showcases the model's ability to maintain consistency and accuracy across complex edits, ensuring that both the forward and reverse transformations align coherently. Additional examples, such as adding and removing objects, further emphasize \model{}'s adaptability in diverse editing tasks. \Cref{fig:reverse_exp} illustrates how our method ensures precise, reversible edits while maintaining the integrity of the original content.

\section{Dataset Details}

\subsection{LLM Prompts for Data Generation}

To ensure reproducibility, we provide the exact prompts used with LLMs for generating reverse instructions and expanding datasets. These prompts were crucial for creating the training data without requiring ground-truth edited images.

\subsubsection{LLM Prompt for Reverse Instruction Generation (IP2P Dataset)}

\begin{small}
\begin{verbatim}
Below is an instruction that describes a 
task, paired with an input that provides 
further context. Write a response that 
appropriately completes the request.

### Instruction:
You are an expert in image editing 
instructions. Given an input caption, an 
edit instruction, and the resulting edited 
caption, generate a reverse instruction 
that would undo the edit and return to 
the original state.

### Input:
Input Caption: "{input_caption}"
Edit Instruction: "{edit_instruction}"
Edited Caption: "{edited_caption}"

Examples:
- Input: "A man wearing a denim jacket" 
  -> "make the jacket a rain coat" 
  -> "A man wearing a rain coat" 
  Reverse: "make the coat a denim jacket"
- Input: "A sofa in the living room" 
  -> "add pillows" -> "A sofa in the 
  living room with pillows"
  Reverse: "remove the pillows"

### Response:
{reverse_instruction}
\end{verbatim}
\end{small}

\subsubsection{LLM Prompt for Multi-Edit Generation (CC3M/CC12M)}

\begin{small}
\begin{verbatim}
Below is an instruction that describes 
a task, paired with an input that 
provides further context. Write a 
response that appropriately completes 
the request.

### Instruction:
Given an input caption, generate 4 
different edit instructions along with 
their corresponding edited captions and 
reverse instructions. Focus on diverse 
edit types: color changes, object 
addition/removal, and positional 
changes. Ensure edits are realistic 
and reversible.

### Input:
Input Caption: "{input_caption}"

### Response:
1. Edit Instruction: [instruction]
   Edited Caption: [result]
   Reverse Instruction: [undo instruction]

2. Edit Instruction: [instruction]
   Edited Caption: [result]
   Reverse Instruction: [undo instruction]

3. Edit Instruction: [instruction]
   Edited Caption: [result]
   Reverse Instruction: [undo instruction]

4. Edit Instruction: [instruction]
   Edited Caption: [result]
   Reverse Instruction: [undo instruction]
\end{verbatim}
\end{small}

\subsection{Dataset Filtering}

We apply CLIP~\cite{Radford2021CLIP} to both the CC3M~\cite{sharma2018conceptual} and CC12M~\cite{changpinyo2021conceptual} datasets to calculate the similarity between captions and images, ensuring that the text descriptions accurately reflect the content of the corresponding images. Following the methodology used in InstructPix2Pix (IP2P)~\cite{Brooks2022InstructPix2Pix}, we adopt a CLIP-based filtering strategy with a similarity threshold set at 0.2. This threshold filters out image-caption pairs that do not have sufficient semantic alignment, allowing us to curate a dataset with higher-quality text-image pairs. For the filtering process, we utilize the CLIP ViT-L/14 model, which provides a robust and well-established framework for capturing semantic similarity across text and images.

By applying this filtering process, we ensure that only relevant and coherent pairs remain in the dataset, improving the quality of training data and helping the model better generalize to real-world editing tasks. As a result, the filtered CC3M dataset contains 2.5 million image-caption pairs, while the filtered CC12M dataset contains 8.5 million pairs. This careful curation of the dataset enhances the reliability of the training process without relying on human annotations, making it scalable for broader real-image datasets without the cost and limitations of human-annotated ground-truth datasets~\cite{Brooks2022InstructPix2Pix, Zhang2023MagicBrush}.

\subsection{More Examples from Reverse Instructions Dataset}

To demonstrate the versatility of our reverse instruction dataset, we provide examples with multiple variations of edits for two different input captions. Each caption has four distinct edits, such as color changes, object additions, object removals, and positional adjustments. This variety helps the model generalize across a wide range of tasks and scenarios, as discussed in Sec. 4.2, main paper. The use of large language models (LLMs) to generate reverse instructions further enhances the flexibility of our dataset.

These examples, along with others in Tab. 1, illustrate the diversity of edit types our model learns, enabling it to perform a wide range of tasks across different real-image datasets. The reverse instruction mechanism ensures that the edits are reversible, maintaining consistency and coherence in both the forward and reverse transformations.

    \section{Additional Implementation Details} \label{subsec:addi_impl}

    \subsection{Details of Competitor Methods}

Our method offers significant advantages over competitors in both training and inference. Unlike supervised methods that rely on paired triplets of input images, edited images, and instructions, our approach eliminates the need for such datasets, reducing biases and improving scalability. For example, MagicBrush is fine-tuned on a human-annotated dataset, while HIVE leverages Prompt-to-Prompt editing with human annotators, introducing dependency on labor-intensive processes. Furthermore, MGIE and SmartEdit rely on LLMs during inference, which significantly increases computational overhead. These distinctions highlight the efficiency and practicality of our approach, as it avoids the need for expensive human annotations and additional inference-time complexities.
Like other editing methods, our approach can produce small variations for different random seeds but consistently applies the specified edit, eliminating the need for manual selection. To the best of our knowledge, the compared methods, \eg{,} MagicBrush, InstructPix2Pix or other methods, also do not involve manual selection.

\vspace{2pt}\noindent\textbf{InstructPix2Pix~\cite{Brooks2022InstructPix2Pix}} is a diffusion-based model that performs instruction-based image editing by training on triplets of input image, instruction, and edited image\footnote{\url{https://github.com/timothybrooks/instruct-pix2pix}}. The model is fine-tuned on a synthetic dataset of edited images generated by combining large language models (LLMs) and Prompt-to-Prompt~\cite{Hertz2022Prompt2Prompt}. This approach relies on paired datasets, which can introduce biases and limit generalization. InstructPix2Pix serves as one of the key baselines for our comparison, given its supervised training methodology.

\vspace{2pt}\noindent\textbf{HIVE~\cite{Zhang2023HIVE}} is an instruction-based editing model that fine-tunes InstructPix2Pix based on human feedback\footnote{\url{https://github.com/salesforce/HIVE}}. Specifically, HIVE learns from user preferences about which edited images are preferred, incorporating this feedback into the model training. While this approach allows HIVE to better align with human expectations, it still builds on top of InstructPix2Pix and does not start training from scratch. This limits its flexibility compared to methods like \model{}, which are trained from the ground up.

\vspace{2pt}\noindent\textbf{MagicBrush~\cite{Zhang2023MagicBrush}} fine-tunes the pre-trained weights of InstructPix2Pix on a human-annotated dataset to improve real-image editing performance\footnote{\url{https://github.com/OSU-NLP-Group/MagicBrush}}. While this fine-tuning approach makes MagicBrush highly effective for specific tasks with ground-truth labels, it limits its generalizability compared to methods like \model{}, which are trained from scratch. Moreover, MagicBrush's reliance on human-annotated data introduces significant scalability challenges, as obtaining such annotations is both costly and labor-intensive. This dependency makes it less suited for broader datasets where large-scale annotations may not be feasible.

\vspace{2pt}\noindent\textbf{MGIE~\cite{fu2023guiding}} introduces a large multimodal language model to generate more precise instructions for image editing\footnote{\url{https://ml-mgie.com/playground.html}}. Like InstructPix2Pix, MGIE requires a paired dataset for training but uses the language model to improve the quality of the instructions during inference. However, this reliance on LLMs during inference adds computational overhead. In contrast, \model{} operates without LLMs at inference time, reducing overhead while maintaining flexibility.

\vspace{2pt}\noindent\textbf{SmartEdit~\cite{huang2024smartedit}} is based on InstructDiffusion, a model already trained for instruction-based image editing tasks\footnote{\url{https://github.com/TencentARC/SmartEdit}}. It introduces a bidirectional interaction module to improve text-image alignment, but its reliance on the pre-trained InstructDiffusion limits flexibility, as SmartEdit does not start training from scratch. Additionally, SmartEdit depends on large language models (LLMs) during inference, increasing computational overhead. This makes SmartEdit less efficient than \model{} in scenarios where real-time or large-scale processing is required.

\vspace{2pt}\noindent\textbf{DiffusionCLIP~\cite{kim2022diffusionclip}} leverages pre-trained diffusion models for text-driven image manipulation by fine-tuning the reverse diffusion process with a CLIP-based loss\footnote{\url{https://github.com/gwang-kim/DiffusionCLIP.git}}. Unlike \model{}, which enforces edit reversibility constraints for unsupervised training, DiffusionCLIP relies on fine-tuning for each new target attribute, making it less scalable for large-scale instruction-based editing. Additionally, its approach requires per-attribute model tuning and inversion of the input image before the editing process, leading to increased training and inference overhead compared to \model{}, which generalizes across a diverse set of edits without explicit supervision and additional inversion process.

During evaluation, we use the publicly available implementations and demo pages of the baseline methods. Each baseline provides a different approach to instruction-based image editing, and together they offer a comprehensive set of methods for comparing the performance, flexibility, and efficiency of the proposed method, \model{}.
    
    \subsection{Code Implementation Overview}
    Our \model{} implementation with ERC builds on existing frameworks for reproducibility:
    
    \begin{itemize}
        \item \textbf{Base Framework:} The code is based on InstructPix2Pix\footnote{\url{https://github.com/timothybrooks/instruct-pix2pix}}, which provides the foundation for instruction-based image editing.
    
        \item \textbf{Adopted CLIP Losses:} We adopted and modified CLIP-based loss functions from StyleGAN-NADA\footnote{\url{https://github.com/rinongal/StyleGAN-nada}} to fit ERC, improving image-text alignment for our specific tasks.
    \end{itemize}
    
    \subsection{Algorithm Overview}
    In this section, we explain the proposed method, \model{}, which introduces unsupervised learning for instruction-based image editing. The core of our approach is the Edit Reversibility Constraint (ERC), which ensures that edits are coherent and reversible when applied sequentially in both forward and reverse instructions.
    
    The algorithm consists of two key processes:
    
    \begin{itemize}
        \item \textbf{Forward Process:} Starting with an input image and a forward edit instruction, noise is first added to the image. The model then predicts the noise, which is applied to reverse the noise process and recover the edited image (\textit{see \cref{alg:\model{}}, lines 2-4}).
        
        \item \textbf{Reverse Process:} Given the forward-edited image and a reverse edit instruction, noise is applied again. The model predicts the reverse noise, which is used to undo the edits and reconstruct the original image. This ensures that the reverse edits are consistent with the original input image (\textit{see \cref{alg:\model{}}, lines 6-8}).
    \end{itemize}

    ERC is applied between the original input image, the forward-edited image, and the reconstructed image, along with their respective attention maps and captions (\textit{see \cref{alg:\model{}}, line 10}). The $\mathcal{L}_{ERC}$ function guides the model’s learning through backpropagation (\textit{see \cref{alg:\model{}}, lines 12-13}).

    \begin{algorithm}
    \caption{Unsupervised Instruction-Based Image Editing (\model{}) with ERC} \label{alg:\model{}}
    \begin{algorithmic}[1]
    \Require Image $I_{i}$ (input image), Forward edit instruction $F$, Reverse edit instruction $R$, Noise levels $t$ (forward), $\hat{t}$ (reverse), Model $M$, Loss function $L_{ERC}$, Noise function $N$, Input caption $T_{i}$, Edited caption $T_{e}$
    \Ensure Edited image $I_{e}$, Reconstructed image $I_{r}$
    
    \vspace{5mm}
    
    \State \textbf{Forward Process:}
    \State $z_t \gets N(I_{i}, t)$ \Comment{Add noise $t$ to the input image $I_{i}$}
    \State $\hat{\epsilon}_F, A_f \gets M(z_t | I_{i}, F)$ \Comment{Model $M$ predicts forward noise $\hat{\epsilon}_F$ and extracts attention map $A_f$}
    \State $I_{e} \gets \text{Apply}(\hat{\epsilon}_F, z_t, t)$ \Comment{Apply predicted noise $\hat{\epsilon}_F$ to reverse the process of obtaining $z_t$ and recover $I_{e}$}
    
    \vspace{5mm}
    
    \State \textbf{Reverse Process:}
    \State $z_{\hat{t}} \gets N(I_{e}, \hat{t})$ \Comment{Add noise $\hat{t}$ to the forward-edited image $I_{e}$}
    \State $\hat{\epsilon}_R, A_r \gets M(z_{\hat{t}} | I_{e}, R)$ \Comment{Model $M$ predicts reverse noise $\hat{\epsilon}_R$ and extracts attention map $A_r$}
    \State $I_{r} \gets \text{Apply}(\hat{\epsilon}_R, z_{\hat{t}}, \hat{t})$ \Comment{Apply predicted noise $\hat{\epsilon}_R$ to reverse the process of obtaining $z_{\hat{t}}$ and recover $I_{r}$}
    
    \vspace{5mm}
    
    \State \textbf{Edit Reversibility Constraint Loss:}
    \State $L_{ERC} \gets L(I_{i}, I_{e}, I_{r}, A_f, A_r, T_{i}, T_{e})$ \Comment{Compute ERC loss using $I_{i}$, $I_{e}$, $I_{r}$, attention maps $A_f$, $A_r$, input text $T_{i}$, and edited text $T_{e}$}
    
    \vspace{5mm}
    
    \State \textbf{Update Model:}
    \State Backpropagate the loss $L_{ERC}$ and update the model $M$
    
    \State Repeat until convergence
    
    \end{algorithmic}
    \end{algorithm}

\section{Limitations and Failure Cases}

While our method demonstrates strong performance across various editing tasks, we acknowledge several limitations. Our reliance on CLIP for semantic alignment introduces challenges in fine-grained spatial reasoning, object counting, and complex compositional understanding. The method may struggle with instructions requiring precise spatial relationships (\eg{,} ``add three apples to the left of the table'') or complex multi-object scenarios with occlusions. Additionally, like many diffusion-based methods, our approach has difficulties with text rendering, extreme lighting changes, and cases requiring simultaneous style transfer with significant structural modifications.

Our method's performance is also constrained by training data quality and the reverse instruction generation process using LLMs. While more efficient than LLM-based approaches during inference, the training process requires additional computational overhead for generating reverse instructions and computing attention consistency losses. Future work could address these limitations by integrating stronger multimodal models for better spatial understanding, incorporating perceptual losses for improved photorealism, and developing more sophisticated evaluation frameworks beyond CLIP-based metrics.

\end{document}